%% file: paper.tex
\documentclass[]{bytedance_seed}

\usepackage[toc,page,header]{appendix}

\usepackage{minitoc}

\title{ KORGym: A Dynamic Game Platform for\\ LLM Reasoning Evaluation  }

\author{ByteDance Seed}
\author{M-A-P}
\author{Beihang University}

\abstract{
 Recent advancements in large language models (LLMs) underscore the need for more comprehensive evaluation methods to accurately assess their reasoning capabilities. Existing benchmarks are often domain-specific and thus cannot fully capture an LLM’s general reasoning potential. To address this limitation, we introduce the \textbf{Knowledge Orthogonal Reasoning Gymnasium (KORGym)}, a dynamic evaluation platform inspired by KOR-Bench~\citep{korbench} and Gymnasium~\citep{gym}. KORGym offers over fifty games in either textual or visual formats and supports interactive, multi-turn assessments with reinforcement learning scenarios. Using KORGym, we conduct extensive experiments on 19 LLMs and 8 VLMs, revealing consistent reasoning patterns within model families and demonstrating the superior performance of closed-source models. Further analysis examines the effects of modality, reasoning strategies, reinforcement learning techniques, and response length on model performance. We expect KORGym to become a valuable resource for advancing LLM reasoning research and developing evaluation methodologies suited to complex, interactive environments.
}

\date{\today}
\correspondence{Jiajun Shi at \email{shijiajun.123@bytedance.com}, Ge Zhang at \email{zhangge.eli@bytedance.com}}

\checkdata[Project Page]{\url{https://github.com/multimodal-art-projection/KORGym}}

\begin{document}
\maketitle

%不需要目录就注释掉 注意目录不要和第一页放在一块 要有\newpage
%\newpage
%\tableofcontents
%\newpage

\input{sections/introduction}
\input{sections/related_work}
\input{sections/approach}
\input{sections/experiments}

\input{sections/analysis}

\input{sections/conclusion}

\clearpage

\bibliographystyle{plainnat}
\bibliography{main}

\clearpage

\beginappendix

\input{sections/appendix}

\end{document}

%% file: sections/introduction.tex
\section{Introduction}

Recent advances in reasoning models have yielded strong performance in tasks such as textual comprehension~\citep{comprehension} and logical inference~\citep{shi2025cryptoxcompositionalreasoning}. However, most benchmarks remain domain-specific (e.g., AIME~\citep{AIME}, PHYBench~\citep{PHYBench}) and fail to capture general reasoning ability. Even benchmarks intended to evaluate broader reasoning (e.g., SuperGPQA~\citep{SuperGPQA}, HLE~\citep{HLE}) are heavily influenced by pretraining data, limiting their capacity to measure intrinsic reasoning skills. To address this gap, we propose a benchmark designed to evaluate the intrinsic reasoning capabilities of LLMs independent of pretraining knowledge. Games, with their diverse scenarios rarely encountered in pretraining corpora, offer an ideal testbed for such evaluation.

While games offer a promising benchmark medium, existing approaches exhibit several shortcomings. LogicGame~\citep{logicgame}, for example, employs only single-turn scenarios, preventing evaluation of long-term planning in LLMs. TextArena~\citep{textarena} and SPINBench~\citep{spin_bench} support multi-turn scenarios but introduce opponent dynamics that generate extraneous variability, confounding pure reasoning assessment and limiting suitability for reinforcement learning (RL) by enabling hacked strategies. Moreover, gg-bench~\citep{ggbench} relies heavily on generative capacity and lacks robustness in both gameplay fidelity and RL integration.

% To address these limitations, we leverage the knowledge-orthogonal reasoning concept of KORBench~\citep{korbench} (see Appendix \ref{kor}) and the reinforcement-learning environment Gymnasium~\citep{gym} to propose the \textbf{Knowledge Orthogonal Reasoning Gymnasium (KORGym)}. KORGym comprises over fifty games across six reasoning dimensions: mathematical and logical reasoning, control interaction reasoning, puzzle reasoning, spatial and geometric reasoning, strategic reasoning, and multimodal reasoning. The platform is structured into four modular components—the Inference Module, the Game Interaction Module, the Evaluation Module, and the Communication Module—which collectively enable multi-round evaluations, configurable difficulty levels, and stable reinforcement-learning support.

%update by jianyang

\begin{wrapfigure}[25]{r}{0.45\textwidth}
  \centering
  \includegraphics[width=0.4\textwidth]{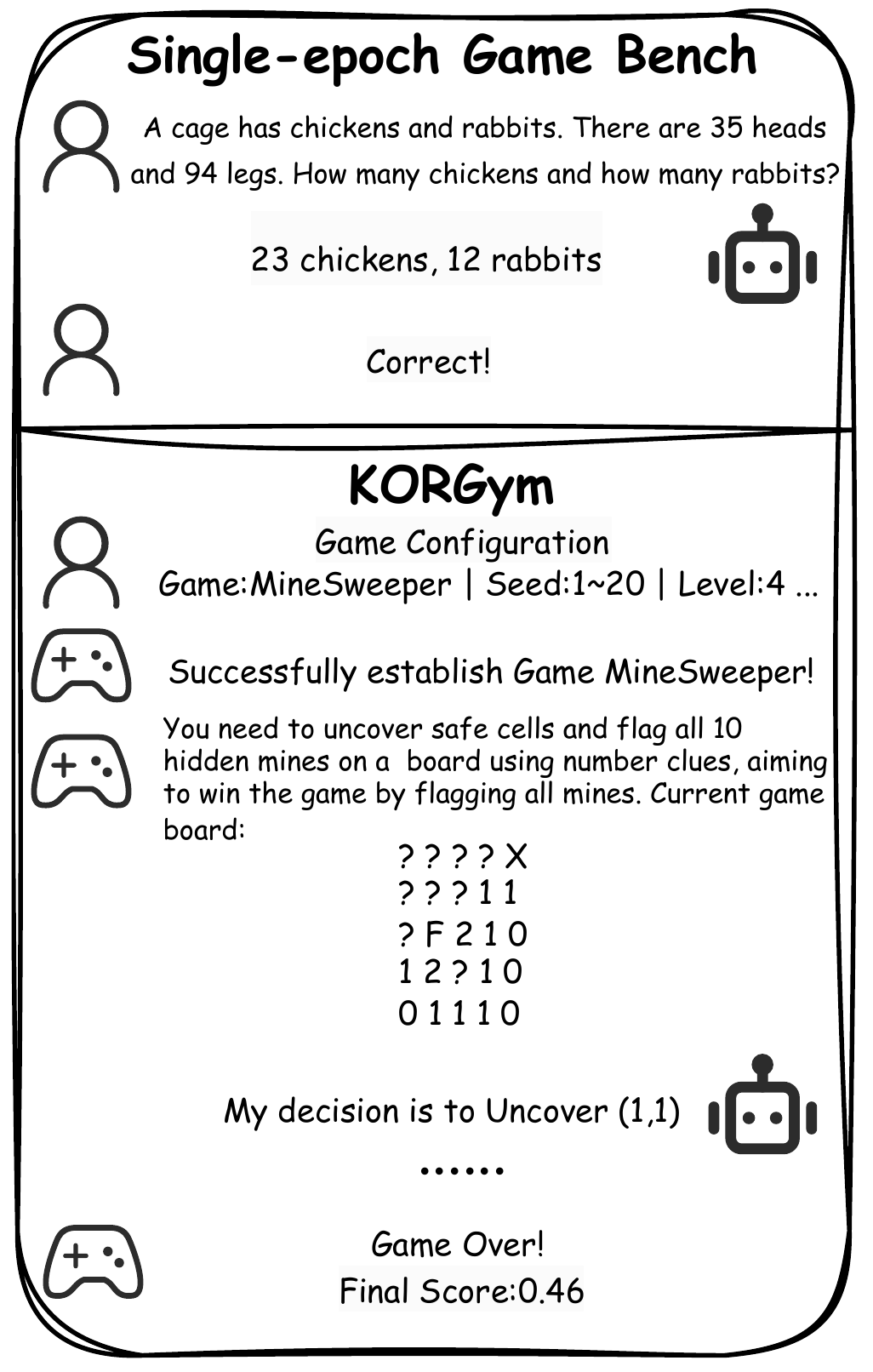}
  \caption{Comparison Between Traditional Single-epoch Game Benchmark and KORGym.}
  % \vspace{-10mm}
  \label{fig:intro}
\end{wrapfigure}

To overcome these limitations, in Figure~\ref{fig:intro}, we introduce the \textbf{K}nowledge \textbf{O}rthogonal \textbf{R}easoning \textbf{Gym}nasium (\textbf{KORGym}), inspired by the knowledge-orthogonal reasoning framework of KORBench~\citep{korbench} (see
Appendix \ref{kor}) and built on the reinforcement-learning environment Gymnasium~\citep{gym}. 

% \begin{figure}
%   \centering
%   \includegraphics[width=0.4\textwidth]{pics/introduction/introduction.pdf}
%   \caption{Comparison Between Traditional Single-epoch Game Benchmark and KORGym.}
%   % \vspace{-10mm}
%   \label{fig:intro}
% \end{figure}

Specifically, KORGym features over fifty games spanning six reasoning dimensions: mathematical and logical reasoning, control interaction reasoning, puzzle reasoning, spatial and geometric reasoning, strategic reasoning, and multimodal reasoning. The platform is organized into four modular components—the inference module, game interaction module, evaluation module, and communication module—enabling multi-round evaluations, configurable difficulty levels, and stable reinforcement-learning support.

By integrating textual and multimodal challenges, KORGym provides a comprehensive assessment of LLMs’ adaptability, strategic planning, and decision-making capabilities, thereby offering a more accurate reflection of their intrinsic reasoning abilities. Using KORGym, we conduct extensive experiments, yielding several key insights:

\begin{itemize}[leftmargin=6mm]
    \item Reasoning abilities display consistent profiles of strengths and weaknesses within the same model series.
    \item Modality influences reasoning performance, with distinct patterns observed between open-source and closed-source LLMs.
    \item Thinking models exhibit behavioral patterns distinct from those of non-thinking models.
    \item LLMs often employ explicit reasoning paradigms during problem-solving, which may partially constrain their performance.
    \item Appropriate reinforcement learning enhances reasoning capabilities and yields more balanced performance across different reasoning dimensions.
\end{itemize}

In summary, our main contributions are as follows:
\begin{itemize}[leftmargin=6mm]
    \item We design a suite of \textbf{over fifty text- and vision-based games} tailored to evaluate the reasoning capabilities of large language models.
    \item We present \textbf{KORGym}, {an extensible framework supporting incremental development and reinforcement-learning integration}.
    \item We conduct a comprehensive empirical analysis of 19 LLMs and 8 vision-language models and uncover several key insights.
\end{itemize}

%% file: sections/related_work.tex
\section{Related Work}
\paragraph{\textbf{LLMs for Gaming.}}
Games serve as valuable testbeds for evaluating large language models (LLMs) due to their demands for multi-step reasoning and strategic planning. Early research focused on single-game evaluations in domains like Minecraft~\citep{mindagent} or social deduction games~\citep{avalonbench,werewolf}, but these narrow settings limited generalizability. Subsequent efforts introduced broader benchmarks with diverse game types and multi-agent frameworks emphasizing coordination or competition, though critical dimensions such as open-ended negotiation, dynamic cooperation-conflict shifts, and rich social dynamics remained underexplored. To address these gaps, SPIN-Bench~\cite{spin_bench} unifies strategic planning and social intelligence by combining formal planning analysis, multi-agent cooperation/competition, and open-ended dialogue. Existing benchmarks vary widely in environment diversity and technical capabilities. Some frameworks offer diverse environments but lack human evaluation, while others focus on specific scenarios yet miss key features. SPIN-Bench stands out with a balanced mix of game types, Gym compatibility, and model vs. model evaluation.

\paragraph{\textbf{Knowledge Orthogonality Based Evaluation.}}
Current AI reasoning benchmarks (e.g., MMLU~\citep{MMLU}, CommonsenseQA~\citep{CommonsenseQA}, MATH~\citep{MATH}) emphasize factual recall and problem-solving but often conflate memorization with reasoning, limiting insight into underlying cognitive processes.
To address this, integration-based benchmarks (e.g., ZebraLogic~\citep{ZebraLogic}, TravelPlanner~\citep{travelplanner}) test adaptability and creativity by requiring pattern recognition, logic, and multi-step reasoning in novel contexts.
While these frameworks advance the focus on contextual problem-solving, they still risk entanglement with domain-specific knowledge biases, as seen in mathematical or logical benchmarks like GSM8K~\citep{GSM8K} and FOLIO~\citep{FOLIO}.
To address these gaps, the concept of knowledge orthogonality advocates decoupling reasoning assessment from prior knowledge and prioritizing rule-following in out-of-distribution scenarios to isolate core abilities such as systematic generalization and hypothesis testing. This paradigm shift—from memorization-driven metrics to knowledge-agnostic, creativity-focused evaluations—establishes a fairer framework for measuring cognitive agility, ensuring models demonstrate genuine reasoning rather than reciting learned patterns and fostering AI systems with robust, human-like adaptability in open-world environments.

%% file: sections/approach.tex
\section{Approach}
%我们提出了一种高效的以游戏为基础的评估LLM进行复杂推理能力的method,可以通过LLM与单轮/多轮，文本/多模态游戏进行交互得到score以体现LLM真实的推理能力。
%我们的系统模块主要分为Inference Module,Game Interation Module,Evaluation Module and Communication Module.首先，用户将Initial Parametres传入系统，由系统进行解析，而后将游戏环境初始化信息传入Game Interation Module,得到游戏环境信息并将其输入到Inference Module得到LLM的action，而后将其传入Game Interation Module,如此进行若干轮次交互后将最后的结果传入Evaluation Module，得到最终分数。其中，初始化参数包括:Game Name,Model Information,Seed,Deploy Port Number,Output Dir
\subsection{Framework}\label{framework}
\begin{figure*}[ht]
\centering
\includegraphics[width=0.85\textwidth]{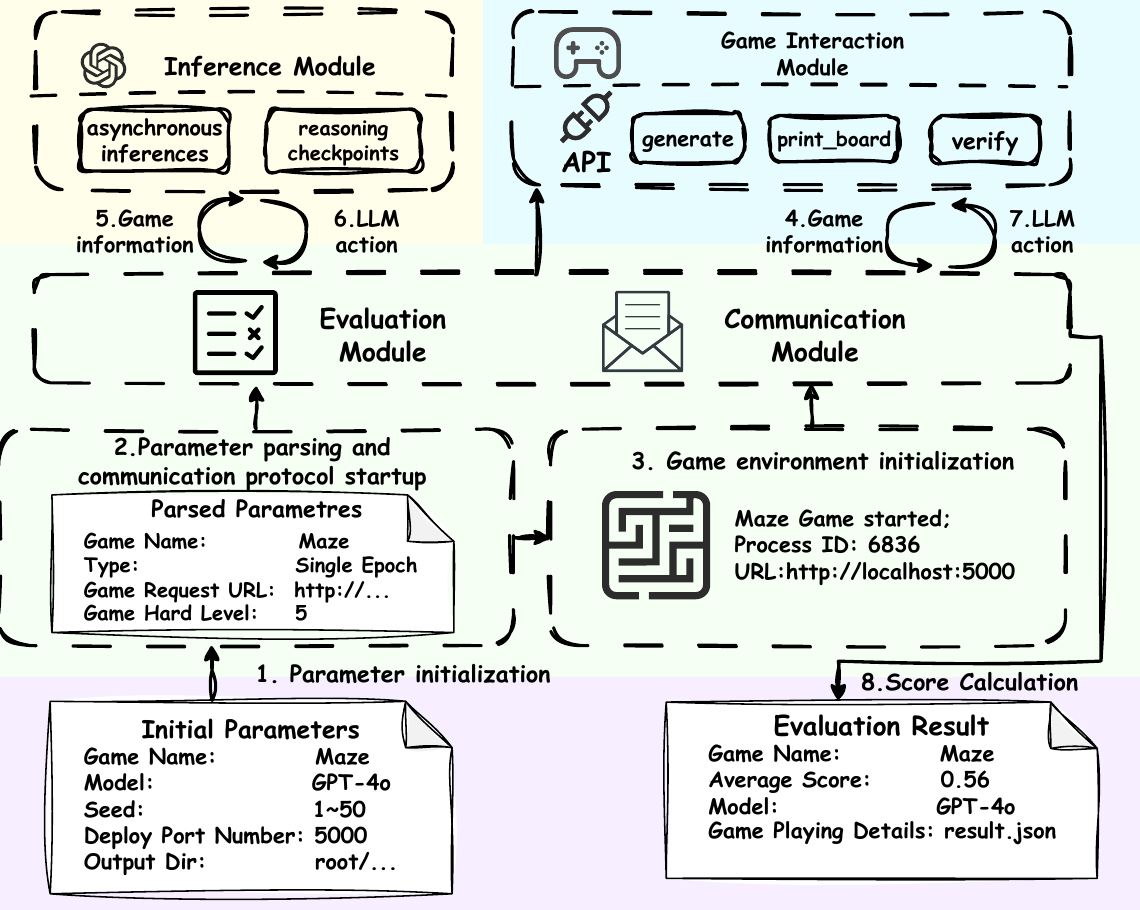}
\caption{Framework of the KORGym system. Our system architecture primarily consists of four modules: the Inference Module, Game Interaction Module, Evaluation Module, and Communication Module.The initialization parameters include: Game Name, Model Information, Seed, Deployment Port Number, and Output Directory.}
\label{fig:system}
\end{figure*}

\input{tables/Overall_tasks}

\begin{figure*}[!htbp]
\centering
\includegraphics[width=\textwidth]{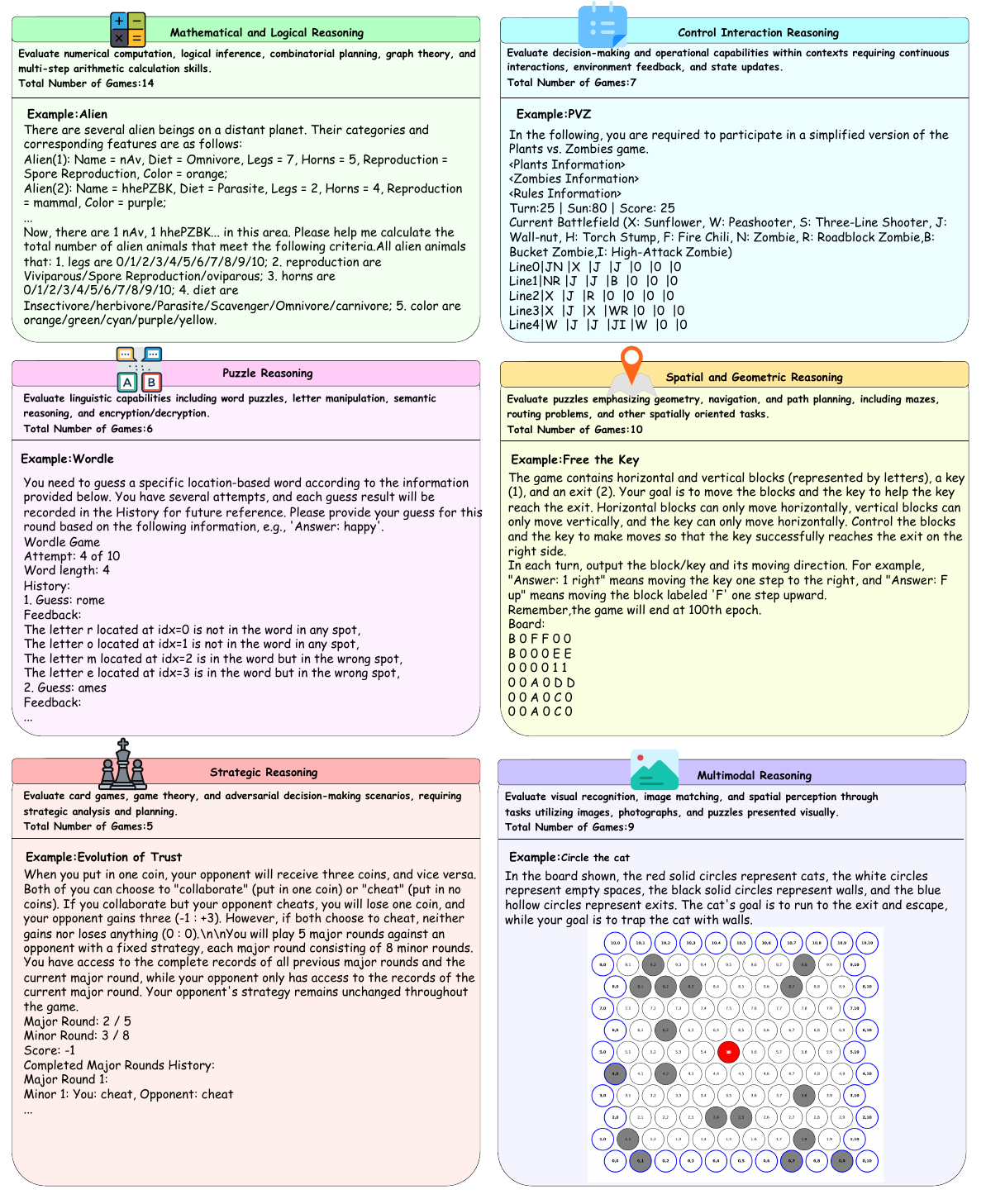}
\caption{Overview of the KORGym tasks. Our KORGym supports over 50 novel games, enabling precise and efficient evaluation of large language models (LLMs) across six distinct capability dimensions.}
\label{fig:overview}
\end{figure*}

We propose \textbf{KORGym}, an efficient game-based framework for evaluating the complex reasoning capabilities of large language models (LLMs) via both single-turn and multi-turn text-based and multimodal games. KORGym is organized into three key modules:

\begin{itemize}[leftmargin=6mm]
    \item \textbf{Evaluation and Communication Module}: the system core, which parses input parameters, establishes inter-module communication protocols, encapsulates and transmits communication packets, and logs final evaluation scores.
    \item \textbf{Game Interaction Module}: encapsulates the game environment and interaction APIs, including:
    \begin{itemize}
        \item \textbf{generate}: initializes the game environment.
        
        \item \textbf{print board}: renders the game board and generates prompts.
        
        \item \textbf{verify}: updates the game state and computes scores.
        
    \end{itemize}
    \item \textbf{Inference Module}: manages model inference processes, including asynchronous acceleration and intermediate result checkpointing.
\end{itemize}

%以上述Module为基础，KORGym的主要推理过程如下：Parameter Initialization:初始参数导入，包括游戏环境名称，游戏环境初始化种子seed等信息；Parameter parsing and communication protocol startup:将传入的初始化参数进行转换，并封装在通信报文单元Item内；Game environment initialization:根据初始化参数对游戏环境进行生成并在Localhost部署服务端；4.获取Game information：Communication Module通过调用generate及print_board获取Game Interation Module生成的游戏信息；5.Game information传递：将获取的游戏信息进行封装，发送给Inference Module的LLM进行推理；6.生成LLM action:从LLM的推理结果中获取LLM action;7.传递LLM action：将提取的LLM action进行封装，而后利用verify API传递到Game Interation Module，进行状态更新及判断；同时，进行游戏状态判断，如果未结束，则跳转到4；否则，跳转到8；8.分数计算及结果输出：通过Evaluation Module计算模型得分，并输出。
Based on these modules, the primary inference workflow of KORGym proceeds as follows:

\begin{itemize}[leftmargin=6mm]
    \item \textbf{Parameter Initialization}: Load initial parameters (e.g., game type, seed, and model).
    \item \textbf{Parameter Parsing and Communication Protocol Startup}: Parse these parameters and encapsulate them into communication packets.
    \item \textbf{Game Environment Initialization}: Generate the game environment according to the parameters.
    \item \textbf{Acquisition of Game Information}: Invoke the generate and print board APIs in the Game Interaction Module to obtain the current environment state.
    \item \textbf{Game Information Transmission}: Package the game environment information for transmission.
    \item \textbf{Generation of LLM Action}: Perform model inference to generate the next action.
    \item \textbf{Transmission of LLM Action}: Send the action via the verify API to update the game state; if the game is not concluded, return to Acquisition of Game Information.
    \item \textbf{Score Calculation and Result Output}: Compute and output the final score.
    
\end{itemize}
%我们的KORGym支持50+个新颖的游戏，并从6个独立的维度来精准高效地评估LLM的推理能力,包括：我们的同期work包括
\subsection{Task Introduction}
\input{tables/comparison}

As illustrated in Figure~\ref{fig:overview}, KORGym supports over fifty novel games, enabling precise and efficient evaluation of the reasoning abilities of large language models (LLMs) across six distinct dimensions (Table~\ref{tab:overall_tasks}): \textbf{Mathematical and Logical Reasoning}, \textbf{Control Interaction Reasoning}, \textbf{Puzzle Reasoning}, \textbf{Spatial and Geometric Reasoning}, \textbf{Strategic Reasoning} and \textbf{Multimodal Reasoning}.

During benchmark development, we select more than fifty games that effectively capture the reasoning capabilities of LLMs. These games span four categories: traditional puzzles (e.g., Sudoku); adaptations of classic video games (e.g., Plants vs. Zombies; Minesweeper); game-theoretic challenges (e.g., N-point; Evolution of Trust); and multimodal tasks (e.g., Jigsaw; Circle the Cat).

%KORGym的相似工作包括\renewcommand{\arraystretch}{1.2}
    % LogicGame~\citep{logicgame}, AgentBench~\citep{AgentBench}, GameArena~\citep{gamearena}, SPIN-Bench~\citep{SpinBench},  TEXTARENA~\citep{textarena} and ReasoningGYM~\citep{reasoninggym}，more detailed comparision can be found in

KORGym offers a suite of over fifty games—with continuous expansion—that support multi-turn interactions via standardized APIs (generate, verify, and print board). The platform is tailored for RL, providing environment states and reward signals, and enables users to adjust game difficulty and environmental diversity through scalable parameters. Additionally, it includes nine multimodal games, facilitating comprehensive evaluation in both textual and multimodal contexts. Related platforms include LogicGame~\citep{logicgame}, AgentBench~\citep{AgentBench}, GameArena~\citep{gamearena}, SPIN-Bench~\citep{SpinBench}, TEXTARENA~\citep{textarena}, and ReasoningGYM~\citep{reasoninggym}. A detailed comparison appears in Table~\ref{tab:comparision}.

%在KORGym的研发阶段，我们

\subsection{Evaluation Method}\label{setting}
%在进行KORGym的评测过程中，由于游戏进度计算的特殊性，故传统的0/1得分计算方法无法有效captureLLM在KORGym上的推理能力，因此我们提出了如下更为全面的得分计算方法：（1）0/1分数制：对于单一目标的游戏，我们采用0分表示失败，1分表示成功，如：7-Maze中以是否达到迷宫出口作为判断标准，38-Free the Key以成功解开谜题使钥匙到达出口为判断标准；（2）correct/total制：对于多个选项进行选择的游戏，我们以选择正确选项数目/总数目作为评分标准，如在44-Jigsaw Puzzle中，我们以摆放正确的拼图数目/总拼图数目作为得分，在32-Minesweeper中，我们以成功标记的地雷/总地雷数作为得分；（3）特定规则下的得分：部分游戏会在达到特定要求后获得相应的分数，并在游戏结束时计算总分数作为得分，如3-2048会在合并相同数字方块时获取对应的score，18-Snake会在吃到苹果时获得1score；
\paragraph{Score Calculation Rules}
To address the limitations of binary (0/1) scoring in reflecting intermediate progress in KORGym, we propose a comprehensive scoring scheme comprising three rules:
\begin{itemize}[leftmargin=6mm]
    \item \textbf{Binary Scoring}: For single-objective games, assign 1 point for success and 0 for failure. For example, in 7-Maze, reaching the exit yields a score of 1.
    \item \textbf{Proportional Scoring}:  For multiple-choice games, the score equals the number of correct responses divided by the total number of options. For instance, in the 44-Jigsaw Puzzle, the score is the number of correctly placed pieces over the total pieces.
    \item \textbf{Cumulative Scoring}: For games that award incremental points, accumulate all points earned. For example, in 3-2048, each tile merge contributes to the final score.
\end{itemize}

\paragraph{\textbf{Capability Dimension Aggregated Mean}}
Raw game scores in KORGym can extend beyond the [0,1] interval and may be skewed by variations in game difficulty or by outlier model behaviors. To mitigate these issues, we introduce \textbf{Capability Dimension Aggregated Mean}, a more robust aggregation metric for evaluating model performance across reasoning dimensions.

Formally, let $G = \{g_1, g_2, \dots, g_N\}$ denote the set of all games,  $M = \{m_1, m_2, \dots, m_K\}$ denote the set of models under evaluation, and  $D = \{d_1, d_2, \dots, d_L\}$represent the set of reasoning capability dimensions. Each game \(g \in G\) is associated with a specific dimension \(d(g) \in D\).
Let \(S_{g,m}\) denote the raw score achieved by model \(m \in M\) on game \(g \in G\).  
For each game \(g\), if the maximum score across all models exceeds 1, i.e., $ \max_{m\in M} S_{g,m} > 1$,we apply a \(\log p\) transformation (i.e., \(\ln(1+x)\)) to compress large score values and reduce skewness; otherwise, we retain the original score:  
\begin{equation} 
S'_{g,m} =
\begin{cases}
\ln\bigl(1 + S_{g,m}\bigr), & \text{if } \displaystyle\max_{m\in M} S_{g,m} > 1,\\[6pt]
S_{g,m}, & \text{otherwise}.
\end{cases}
\end{equation}
To normalize scores across games, we further define, for each game \(g\),
\begin{equation}
a_g = \min_{m\in M} S'_{g,m}, \quad b_g = \max_{m\in M} S'_{g,m}.
\end{equation} 
If \(b_g = a_g\), meaning all models perform identically on game \(g\), we assign every model a normalized score of 0.5 to avoid division by zero. Otherwise, we normalize its adjusted score:
\begin{equation}
\widetilde{S}_{g,m} = \frac{S'_{g,m} - a_g}{b_g - a_g}, \quad \forall m \in M.
\end{equation}
This normalization ensures that for each game, model performances are mapped into the \([0,1]\) range while preserving relative differences.
Subsequently, for each capability dimension \(d \in D\), we define the corresponding set of games and aggregated score of model \(m\) on dimension \(d\) as:
\begin{equation}
G_d = \{g \in G : d(g) = d\},\overline{S}_{d,m} = \frac{1}{|G_d|} \sum_{g\in G_d} \widetilde{S}_{g,m}.
\end{equation}
The resulting matrix \(\{\overline{S}_{d,m}\}_{d\in D,\;m\in M}\) provides a normalized, dimension-wise evaluation of reasoning capabilities that is fair across heterogeneous games.

%% file: tables/Overall_tasks.tex
\begin{table}[!ht]
    \centering
    \small
    \caption{Game Introduction of KORGym. From top to bottom, the categories are: Mathematical and Logical Reasoning games, Puzzle Reasoning games, Spatial and Geometric Reasoning games, Strategic Reasoning games, Control and Interaction Reasoning games, and Multimodal Reasoning games.}
    \resizebox{0.99\textwidth}{!}{
    \begin{tabular}{ll}
        \toprule
        Name & Task Content\\
        \midrule
         Crossword Puzzle        & This game challenges LLM to infer current date given a future date and number of days between them.\\
        Sudoku & This game evaluates an LLM's logical reasoning by solving a Sudoku puzzle.\\
         Light Out Game & This game tests an LLM's strategic reasoning by requiring it to switch off all lights on a 3x3 grid.\\
         Square Addition & This game requires LLM to compute column sums based on symbolic values. \\
         Alien & This game requires LLM to count alien species based on multiple traits from a complex dataset.\\
         Party Time & This game challenges LLM to identify and count students who meet specific criteria.\\
         Path Planning Problem & This game requires LLM to calculate the shortest distance between two cities within a complex network.\\
         Construction Company & This game challenges LLM to calculate the minimum time across multiple companies and projects.\\
         Tower of Hanoi & This game requires LLM to solve a Tower of Hanoi puzzle, moving disks between columns to reach a goal state.\\
         Numeric Bricks & This game challenges an LLM to fill a grid by expanding each labeled cell according to a specified count.\\
         One Stroke Drawing & This game requires LLM to find an Eulerian path that visits every edge exactly once.\\
         Nullify & This game challenges an LLM to combine arithmetic units using operations to achieve a final result of zero.\\
         Coloring Issue & This game challenges an LLM to assign colors to graph nodes such that no two adjacent nodes share the same color.\\
         City Traveller &This game requires LLM to extract, filter, and compute city information from a complex city network.\\
         \midrule
         Word Problem & This game challenges an LLM to find a specific English word that matches several constraints.\\
        Alphabetical Sorting & This game requires LLM to rearrange the remaining unordered letters to form a valid word. \\
        Letter Connection & This game requires LLM to reconstruct a hidden word from a 3x3 letter grid.\\
        Word Transformation & This game requires LLM to decode a transformed word by reasoning through layered transformations .\\
        Wordle & This game requires LLM to perform deductive reasoning through iterative word guessing based on structured feedback.\\
        Crypto Word & This game requires LLM to decode a sentence by mapping emojis to letters through iterative feedback-based guessing.\\
        \midrule
        Maze & This game requires LLM to find a valid path through a maze from the start point to the destination. \\
        Sokoban & This game requires LLM to solve a Sokoban puzzle, pushing all boxes onto the target areas. \\
        Playlines & This game requires LLM to fill in grids by connecting identical numbers on a grid without leaving empty spaces .\\
        Emoji Connect & This game requires LLM to count repeated horizontal or vertical patterns in emoji grids.\\
        8-puzzle & This game requires LLM to plan valid tile moves to reposition a target tile in a sliding puzzle grid.\\
        Bubble Ball Sorting & This game requires LLM to sort colored balls into uniform tubes under stacking constraints.\\
        Pipe Game & This game requires LLM capability to rotate pipes in a grid to create a continuous path.\\
        Free the Key & This game requires LLM to move the building blocks and keys to reach the exit.\\
        Map Simulation & This game tests an LLM’s ability to simulate multi-step movement through a dynamic grid.\\
        Arrow-pathway & This game tests an LLM’s ability to navigate a maze by sequencing directional actions to trigger waypoints.\\
        \midrule
        2048 & This game requires LLM to play the 2048 puzzle by choosing the best move based on the current game board \\
        Trust Evolution & This game requires LLM to identify and exploit opponent behavior patterns through strategic decision-making .\\
        N-point & This game requires LLM  to play expanded 21-Point in dynamic thresholds and an opponent’s fixed behavior.\\
        Spider Solitaire & This game tests an LLM’s ability to plan and execute strategic moves in Spider Solitaire.\\
        Circle the Cat & This game requires LLM to strategically place walls on a hexagonal grid to trap a moving cat.\\
        \midrule
         Minigrid &This game requires LLM to solve a series of tasks based on the Minigrid \citep{MinigridMiniworld23} reinforcement learning system.  \\
        Snake & This game requires LLM to control a growing snake on a bounded grid, avoiding collisions to maximize score.\\
        Tetris & This requires LLM to plan by strategically rotating and placing Tetris blocks to clear rows and maximize score.\\
        Minesweeper & This game requires  LLM to uncover safe cells and flag hidden mines .\\
        PVZ & This game requires LLM to place plants to counter increasingly strong zombies.\\
        Long Cat & This game requires LLM to plan efficient movement sequences by navigating a sliding cat to fill all empty spaces.\\
        Black White Copy & This game requires LLM to toggle rows to transform the board into a specified black-and-white target pattern.\\
        \midrule
        Crossword Puzzle & This game requires LLM to solve linguistic clues to fill the grid with words correctly. \\
        Jigsaw Puzzle & This game requires LLM to match visual puzzle pieces with numbered slots .\\
        Find The Pattern & This game requires LLM to identify the correct visual piece that completes a given pattern.\\
        Circle The Cat (Visual) & This game requires LLM to analyze a visual board and determine optimal wall placements to prevent a cat.\\
        Map Simulation (Visual) & This game requires LLM to interact with diverse objects, and accurately calculate the final position.\\
        Sokoban (Visual) & This game requires LLM to interpret a visual Sokoban puzzle and generate a precise series of moves.\\
        Bubble Ball Sorting (Visual) & This game requires LLM to generate valid moves to achieve uniform color sorting across specified tubes.\\
        Wordle (Visual) & This game requires LLM to deduce the correct secret word through multiple turns of guessing.\\
        Square Addition (Visual) & This game requires LLM to infer integer values to compute accurate column sums.\\
        \bottomrule
    \end{tabular}
    }
    \label{tab:overall_tasks}
\end{table}

%% file: tables/comparison.tex
\begin{table}[h]
    \centering
    \small
    % \renewcommand{\arraystretch}{1.2}
    % LogicGame~\citep{logicgame}, AgentBench~\citep{AgentBench}, GameArena~\citep{gamearena}, SPIN-Bench~\citep{SpinBench},  TEXTARENA~\citep{textarena} and ReasoningGYM~\citep{reasoninggym}
    \caption{Comparison of different game benchmarks.}
    \begin{tabular}{lccccc}
        \toprule
        Benchmark & \# Games & Multi-turn & RL & Controllable Difficulty & Multimodal\\
        \midrule
        LogicGame~\citep{logicgame} & 27 &  \ding{55} & \ding{55} & \ding{55} & \ding{55}\\
        AgentBench~\citep{AgentBench} & 8 & \ding{51} & \ding{55} & \ding{55} & \ding{55}\\
        GameArena~\citep{gamearena} & 3 & \ding{51} & \ding{55} & \ding{55} & \ding{55}\\
        SPIN-Bench~\citep{SpinBench} & 6 & \ding{51} & \ding{51} & \ding{55} & \ding{55}\\
        TEXTARENA~\citep{textarena}& 70+ & \ding{51} & \ding{51} & \ding{51} & \ding{55}\\
        ReasoningGYM~\citep{reasoninggym}& 100+ & \ding{55} & \ding{51} & \ding{51} & \ding{55}\\
        KORGym & 51 & \ding{51} & \ding{51} & \ding{51} & \ding{51}\\
        \bottomrule
    \end{tabular}
    % }
    
    \label{tab:comparision}
\end{table}

%% file: sections/experiments.tex
\section{Experiments}
\subsection{Settings}\label{setting}

%为了充分测试LLM在KORGym的表现，我们对主流的开源和闭源LLM进行了评估，包括
\paragraph{\textbf{LLMs}} 
To comprehensively evaluate LLM performance, we assessed 19 large language models—including 11 thinking models and 8 instruction-tuned models—and 8 vision-language models (Table~\ref{tab:models}).

\begin{table}[!ht]
    \centering
    \small
    \caption{Models evaluated in KORGym.}
    \resizebox{\textwidth}{!}{
    \begin{tabular}{ll}
        \toprule
        Series & Model\\
        \midrule
         GPT & GPT-4o~\citep{openai2024gpt4o}, O1~\citep{o1}, O3-mini\citep{o3}\\
         Claude & Claude-3.7-Sonnet~\citep{claude3.7},Claude-3.7-Sonnet-thinking\\
         Doubao & Doubao-1.5-pro~\citep{doubao1.5pro}, Doubao-1.5-thinking-pro~\citep{doubao-thinking-pro}, Doubao-vision-pro~\citep{Doubao-vision-pro}\\
         DeepSeek & DeepSeek-v3~\citep{deepseekv3}, DeepSeek-R1~\cite{guo2025deepseek}, DeepSeek-R1-Distill-Qwen(7B,32B)\\
         Qwen & Qwen3-32B~\citep{qwen3}, Qwen2.5-Instruct(7B,32B,72B)~\citep{qwen2.5}, Qwen-QwQ-32B~\citep{qwenqwq}, Qwen2.5-VL-Instruct(7B,32B,72B)~\citep{qwen2.5-VL}\\
         Gemini & Gemini-2.0-Flash~\citep{gemini-thinking}, Gemini-2.0-Flash-thinking,Gemini-2.5-pro~\citep{gemini2.5}\\
         InternVL & InternVL3-68B~\citep{internvl3}\\
        \bottomrule
    \end{tabular}
    }
    
    \label{tab:models}
\end{table}

\input{tables/exp/overall}
\input{tables/exp/multimodal}
\paragraph{\textbf{Evaluation Setting}} 

During evaluation, we apply distinct protocols for single-epoch and multiple-epoch games:

\begin{itemize}[leftmargin=6mm]
    \item Single-epoch Games: Each model is evaluated on 50 independently initialized game instances by varying the ``seed'' parameter in the ``generate'' API from 1 to 50.
    \item Multiple-epoch Games: For each model, we initialize 20 game environments. Each episode permits up to 100 interaction rounds, and we vary the ``seed'' parameter in the ``generate'' API from 1 to 50 for reproducibility.
\end{itemize}
 
All assessments use a zero-shot prompting setup to gauge genuine reasoning capabilities, retaining each model’s default sampling parameters (temperature and top-p). We evaluate closed-source models via their hosted APIs and open-source models on eight NVIDIA A100-80G GPUs.

\subsection{Main Results}\label{result}

Table~\ref{tab:overall_performance} reports the performance of LLMs and VLMs on KORGym using the Capability Dimension Aggregated Mean (Section~\ref{setting}). Table~\ref{tab:multimodal_game} presents VLM performance on multimodal tasks and detailed raw scores appear in Appendix~\ref{Appendix:detailed_scores}. The leaderboard will be updated on GitHub after submission. Across 51 games and six reasoning dimensions, the normalized scores yield the following insights.

% 在文中插入双子图
\begin{figure}[!htbp]
  \centering
  % 第一张子图
  \begin{subfigure}[b]{0.5\textwidth}
    \includegraphics[width=\linewidth]{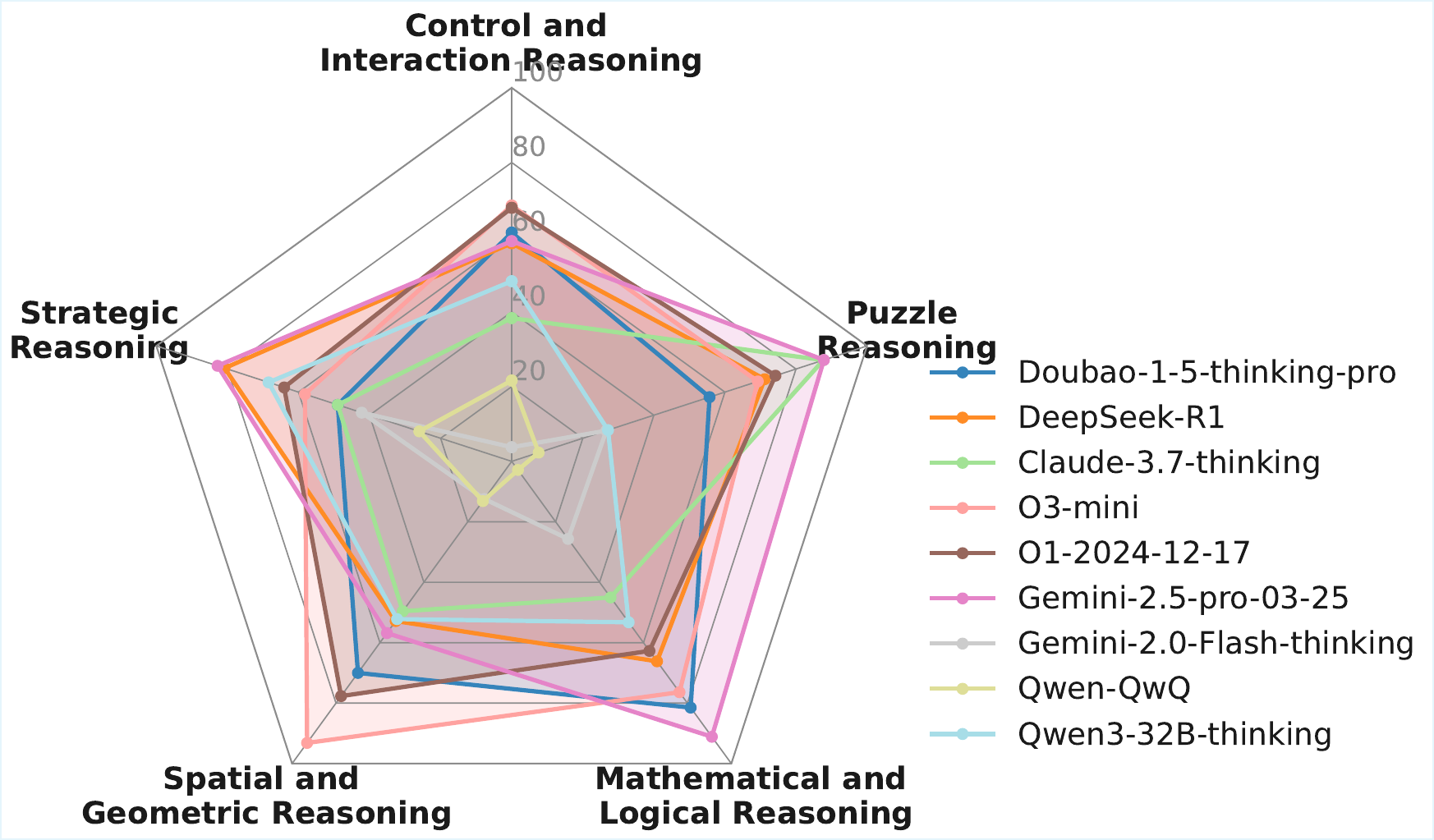}% 吞掉行尾空白
    \caption{Model Ability Analysis}%      
    \label{fig:ability_analysis}%
  \end{subfigure}%
  % 第二张子图
  \begin{subfigure}[b]{0.5\textwidth}
    \includegraphics[width=\linewidth]{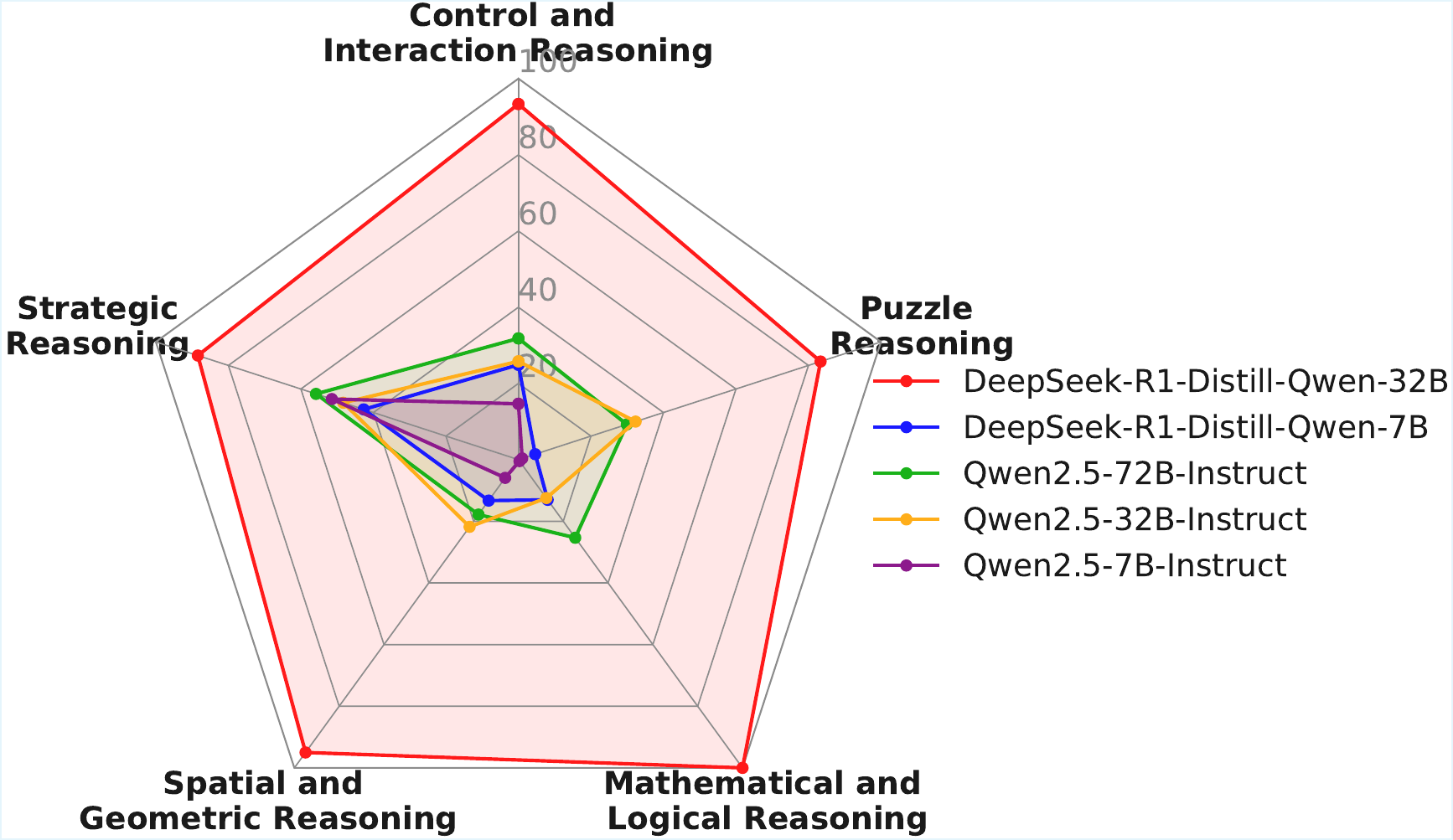}%
    \caption{Model Scale and Architecture Analysis}%
    \label{fig:scale_analysis}%
  \end{subfigure}%
  \caption{Capability Dimension Illustration. Figure (a) showcases the performance of the top-performing models on KORGym. Figure (b) showcases the impact of model scale and architecture on reasoning capabilities.}
  \label{fig:dimension_analysis}
\end{figure}

\paragraph{\textbf{Similar Strength–Weakness Profiles Within Same Model Series}}
Figure~\ref{fig:ability_analysis} shows that O1 and O3-mini excel in spatial reasoning, whereas the Gemini series leads in mathematical and puzzle reasoning.
\paragraph{\textbf{Closed-Source Models Demonstrate Superior Reasoning Performance}}
O3-mini achieves the highest overall score on KORGym, particularly in spatial reasoning. Claude-3.7-thinking and Gemini-2.5-pro top puzzle reasoning, while Doubao-1.5-thinking-pro and DeepSeek-R1 deliver balanced performance across dimensions. In contrast, open-source models lag behind.

\paragraph{\textbf{Impact of Model Scale and Architecture on Reasoning Capabilities}}
Figure~\ref{fig:scale_analysis} demonstrates that model performance scales positively with model size and thinking models outperform size-matched non-thinking variants. For instance, DeepSeek-R1-Distill-Qwen-32B, though smaller in scale, exceeds the performance of Qwen2.5-72B-Instruct.

%% file: tables/exp/overall.tex
\begin{table}[h]
    \centering
    \small
    \caption{Overall performances of different models on KORGym. Model capability dimensions include Mathematical and Logical Reasoning (MLR), Control Interaction Reasoning (CIR), Puzzle Reasoning (PR), Spatial and Geometric Reasoning (SGR) and Strategic Reasoning(SR).}
    \resizebox{0.9\textwidth}{!}{
    \begin{tabular}{lccccccc}
        \toprule
        Model  & \makecell{MLR(\%)} 
  & \makecell{CIR(\%)} 
  & \makecell{PR(\%)} 
  & \makecell{SGR(\%)}  
  & \makecell{SR(\%)}
  & \makecell{Avg.(\%)}\\
 
        \midrule
    O3-mini                  & 77 & 81 & 79 & 94 & 76 &  82  \\
Gemini-2.5-pro-03-25     & 63 & 94 & 93 & 59 & 84 &  79 \\
O1-2024-12-17            & 74 & 83 & 65 & 79 & 66 &  73 \\
Doubao-1-5-thinking-pro  & 65 & 74 & 84 & 72 & 65 &  72\\
DeepSeek-R1              & 66 & 82 & 69 & 56 & 83 &  71\\
Claude-3.7-thinking      & 50 & 93 & 52 & 53 & 64 &  62\\
Qwen3-32B-thinking       & 58 & 55 & 58 & 55 & 71 &  60\\
DeepSeek-v3-0324         & 35 & 55 & 27 & 26 & 69 &  42\\
DeepSeek-R1-Distill-Qwen-32B & 45 & 28 & 35 & 33 & 56 & 39\\
Gemini-2.0-Flash-thinking & 25 & 53 & 34 & 18 & 58 &  38\\
Claude-3.7               & 25 & 55 & 26 & 17 & 50 &  35\\
Qwen-QwQ                 & 37 & 39 & 14 & 18 & 33 &  28\\
Gemini-2.0-Flash         & 24 & 28 & 17 & 12 & 51 &  26\\
GPT-4o                   & 12 & 25 & 8 & 11 & 53 &  22\\
Doubao-1.5-pro           & 18 & 16 & 16 & 7 & 44 & 20\\
Qwen2.5-72B-Instruct     & 18 & 10 & 4 & 7 & 49 & 18\\
Qwen2.5-32B-Instruct     & 13 & 7 & 4 & 9 & 46 & 16\\
DeepSeek-R1-Distill-Qwen-7B & 10 & 2 & 6 & 3 & 33 & 11\\
Qwen2.5-7B-Instruct      & 7 & 1 & 1 & 1 & 29 & 8\\

        \bottomrule
    \end{tabular}
    }
    
    \label{tab:overall_performance}
\end{table}

%% file: tables/exp/multimodal.tex
\begin{table}[h]
    \centering
    \small
    \caption{Multimodal reasoning abilities  of different models on KORGym.}
    \resizebox{0.99\textwidth}{!}{
    \begin{tabular}{lccccccccccc}
        \toprule
        Model  & \makecell{Crossword\\Puzzle(\%)} 
  & \makecell{Jigsaw\\Puzzle(\%)} 
  & \makecell{Find The\\Pattern(\%)} 
  & \makecell{Circle The\\Cat(\%)}  
  & Map Simulation  (\%)
  & Sokoban (\%)
  & \makecell{Bubble Ball\\Sorting(\%)}  
  & Wordle  (\%)
  & \makecell{Square\\Addition(\%)}  \\
        \midrule
Doubao-vision-250115     & 14.4 & 12.9 & 42 & 0 & 2 & 4 & 0 & 0 & 0 \\
Gemini-2.5-Pro            & 18.4 & 19.5 &   66   & 15 & 26& 10 & 90 & 85 & 2 \\
Gemini-2.0-Flash          & 24.9 & 12.7 & 54 & 5 & 0 & 4 & 45 & 15 & 0 \\
GPT-4o                    & 23.7 & 8.4 & 36 & 0 & 0 & 4 & 65 & 15 & 0 \\
Qwen2.5VL-72B            & 14.4 & 10.9 &   28   & 0 & 2 & 4 & 20 & 5 & 0 \\
Qwen2.5VL-32B            & 6.4 & 8.6 &   28   & 0 & 0 & 2 & 25 & 5 & 0 \\
Qwen2.5VL-7B             & 4.4 & 0 &   16   & 0 & 0& 0 & 0 & 0 & 0 \\
InternVL3-78B            & 11.8 & 10 &   38   & 0 & 2 & 0 & 25 & 0 & 0 \\
        \bottomrule
    \end{tabular}
    }
    
    \label{tab:multimodal_game}
\end{table}

%% file: sections/analysis.tex
\section{Discussion}
\subsection{RQ1: Does Modality Affect Reasoning Performance?}

%相同模型在同一游戏的文字版本和多模态版本上的表现，其中，柱状图颜色代表游戏，柱状图是否含阴影代表text/visual版本，折线图实线/虚线代表text/visual版本的平均分数。

\begin{figure*}[ht]
\centering
  \includegraphics[width=0.9\linewidth]{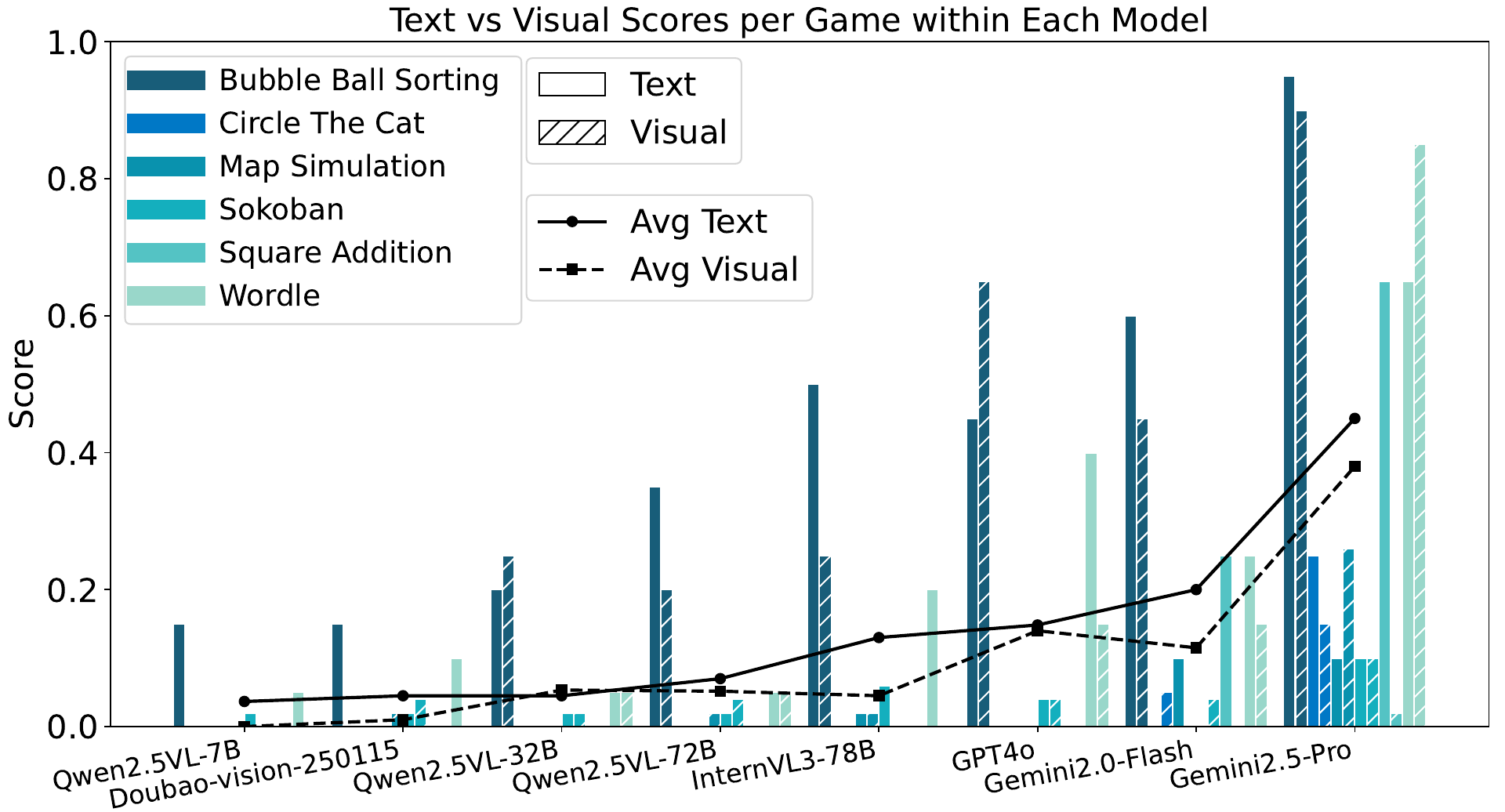}
  \caption{Performance Comparison Between Textual and Multimodal Game Versions. This figure illustrates a given model’s performance on both the textual and multimodal versions of the same game. Different games are represented by distinct bar colors, and bar shading differentiates text (unshaded) from visual (shaded) versions. Solid and dashed lines correspond to the average textual and visual scores, respectively.}
  \label{fig:visual_analysis}
\end{figure*}

%Textual-version Game vs Multimodal-version Game.我们选取4个闭源模型和4个开源模型作为待测模型，测试了6个游戏的text和visual版本，得出了如下结论：1.整体上，文本版本游戏均分大于visual版本游戏均分；2.开源VLM模型text推理能力往往优于visual模型推理能力；3。闭源VLM部分visual游戏得分高于其文字版本对应得分；4.对于数学类游戏，明显文字版本得分高于visual版本得分
\paragraph{\textbf{Textual-version Game vs. Multimodal-version Game.}} As shown in Figure ~\ref{fig:visual_analysis} compares the performance of closed-source and open-source VLLMs on textual and visual versions of six representative games.  Our key findings are:
\begin{itemize}[leftmargin=6mm]
    \item \textbf{Average scores on textual versions consistently exceed those on visual versions. }
    \item \textbf{Open-source VLMs perform better on text-based reasoning than on visual-based tasks}, indicating limited visual grounding or underdeveloped multimodal alignment.
    \item \textbf{Some closed-source VLMs score higher on visual versions than on textual versions}, suggesting stronger visual reasoning or superior multimodal integration.
    \item \textbf{In mathematics-related games, models score significantly higher on textual versions,} highlighting the advantage of symbolic representation for numerical reasoning.
\end{itemize}

% \begin{wrapfigure}[18]{r}{0.55\textwidth}
%   % 可选参数 [n] 指定环绕图片的文本行数，比如 [14]
%   % \begin{wrapfigure}[14]{r}{0.5\textwidth}
%   \centering
%   \includegraphics[width=0.9\linewidth]{pics/exp/text_vs_visual.pdf}
%   \caption{Performance Comparison Between Textual and Multimodal Game Versions. This figure illustrates a given model’s performance on both the textual and multimodal versions of the same game. Different games are represented by distinct bar colors, and bar shading differentiates text (unshaded) from visual (shaded) versions. Solid and dashed lines correspond to the average textual and visual scores, respectively.}
%   \label{fig:visual_analysis}
% \end{wrapfigure}

\subsection{RQ2: Do Different Model Series Exhibit Consistent Behavioral Patterns?}

\begin{figure}[h]
\vspace{-5mm}
  \centering
  % 第一张子图
  \begin{subfigure}[b]{0.49\textwidth}
    \includegraphics[width=\linewidth]{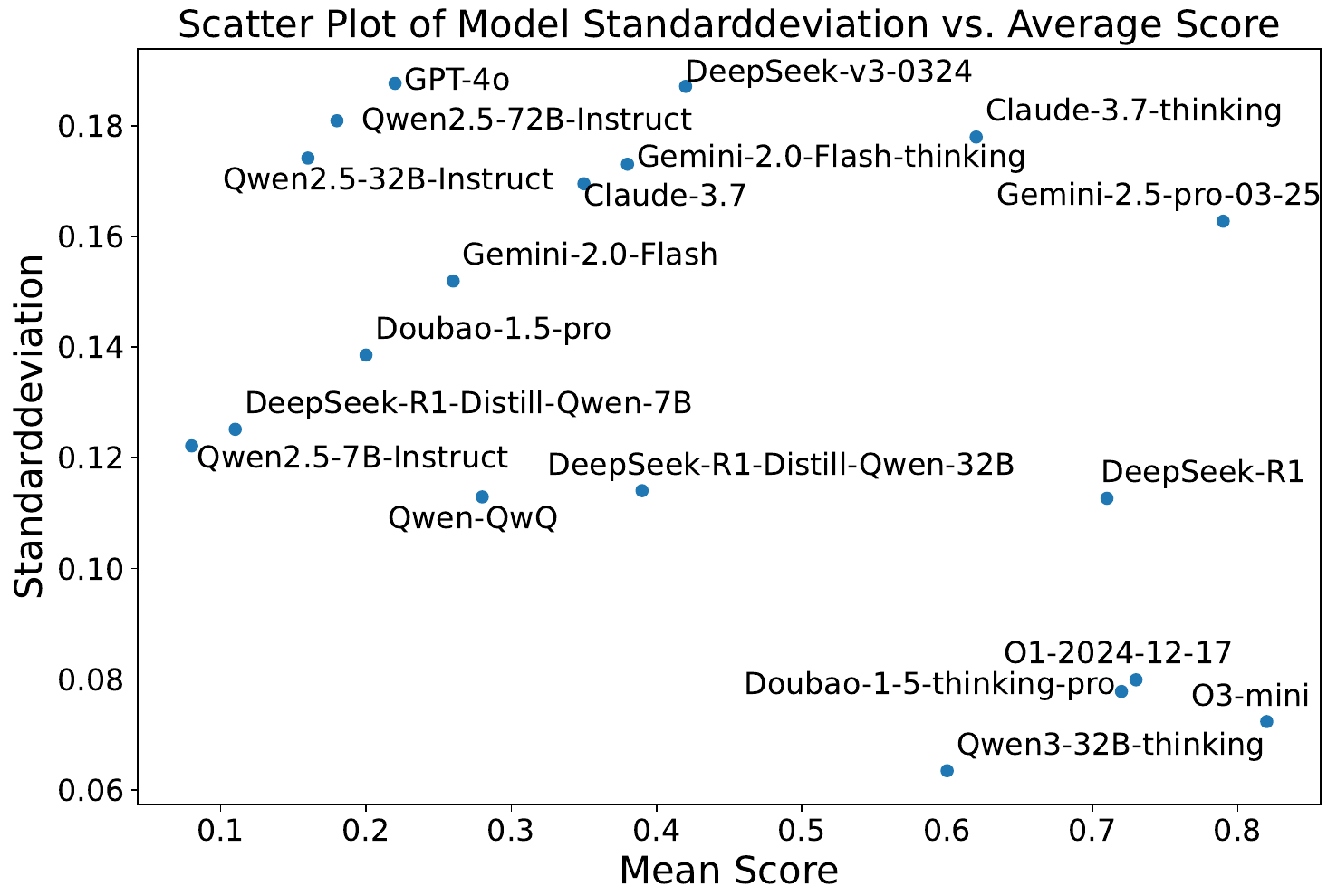}% 吞掉行尾空白
    \caption{Stability Analysis of Model Reasoning Capabilities}%      
    \label{fig:stability}%
  \end{subfigure}%
  % 第二张子图
  \begin{subfigure}[b]{0.5\textwidth}
    \includegraphics[width=\linewidth]{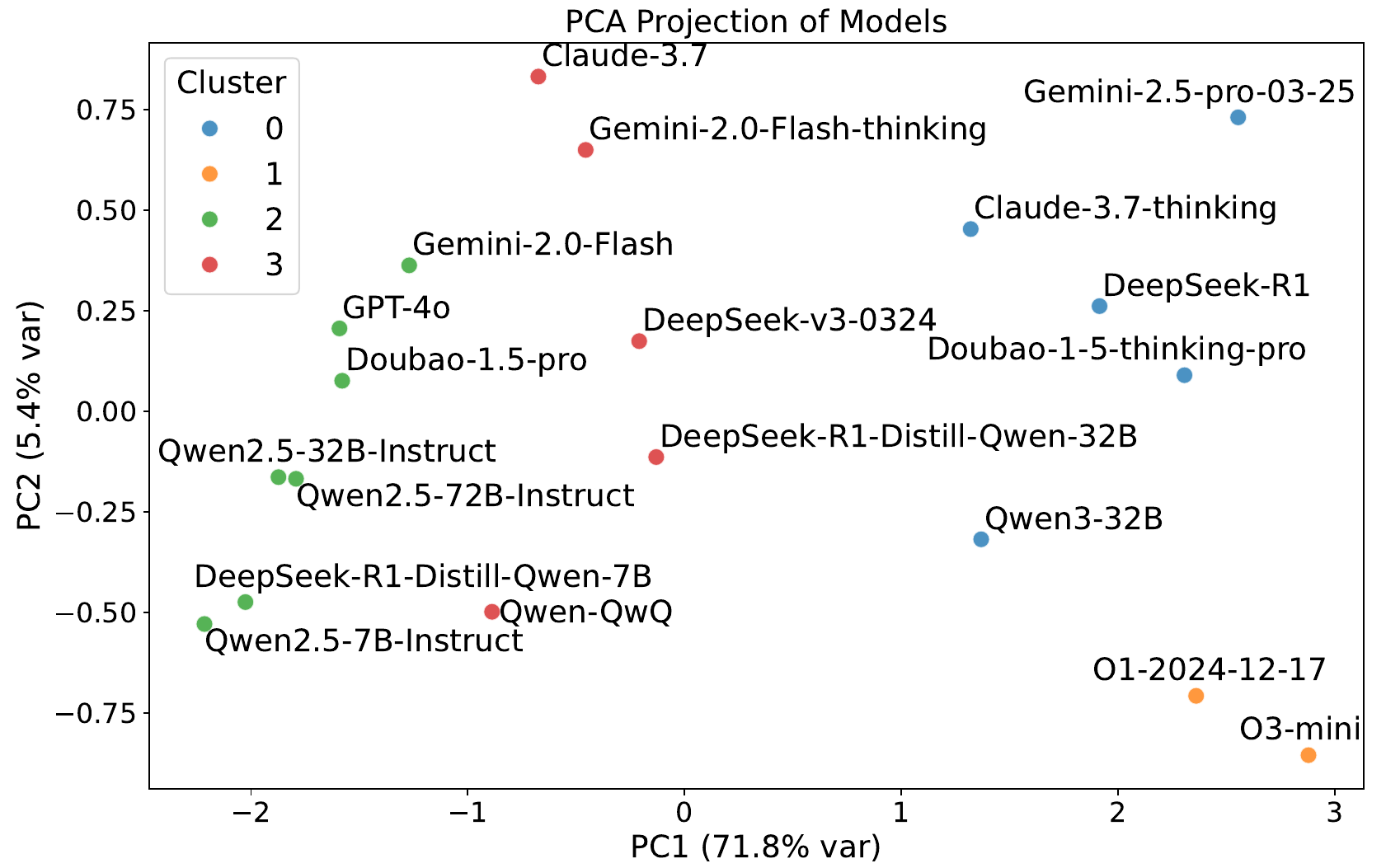}%
    \caption{PCA Projection of Models}%
    \label{fig:PCA}%
  \end{subfigure}%
  \caption{Analysis of Model Behavioral Characteristics. (a) illustrates the relationship between model stability and reasoning performance, whereas (b) depicts the clustering of models based on behavioral traits.}
  \label{fig:dimension_analysis}
\end{figure}

Based on the experimental scores, we computed the mean and standard deviation of each dimension’s scores (Figure \ref{fig:stability}) and performed principal component analysis (PCA) on the score matrix $\hat{S}\in R^ {M\times G}$, where $M$ and $G$ denote the numbers of models and games, respectively (Figure \ref{fig:PCA}). These analyses reveal dominant patterns in the models’ reasoning behavior across five dimensions.

%top level的模型表现出相似的行为特征:top level的模型,如O1,O3-mini往往聚类分布接近，且兼顾较强的推理能力和且无明显能力短板；而次级的模型，如Claude-3.7-thinking,Qwen3往往在推理能力上存在不均衡的问题；
%thinking model与非thinking model行为模式差别明显：在PCA分析中，前两层级全部为推理模型，第四层级几乎全部为非推理模型；
\paragraph{\textbf{Top-Tier Models Exhibit Homogeneous Behavioral Profiles}}
In PCA space, top-tier models (e.g., O1 and O3-mini) form tight clusters, indicating consistently strong reasoning performance across all dimensions. By contrast, secondary models (e.g., Claude-3.7-thinking and Qwen3) display imbalanced performance across reasoning dimensions.
\paragraph{\textbf{Distinct Behavioral Patterns Between Thinking and Non-Thinking Models}} PCA reveals that the first two clusters consist exclusively of thinking models, whereas the fourth cluster comprises almost solely non-thinking models.

% 在文中插入双子图

%由此可见，Cluster 1为GPT-o系列推理模型，在思维模式上趋于一致；而Cluster 0则包括目前表现顶级的推理模型，且Claude-3.7,Gemini-2.5 DeepSeek-R1表现更为一致；Qwen3和Doubao-1.5-thinking-pro则趋于Cluster 1和Cluster 0的边界，说明其与O系列推理模型以及上述三个推理模型均有相似之处；而Cluster 3则包含了部分开源推理模型以及闭源非推理模型，推理能力处于中间水平；Cluster 2则包括了目前主流的开源非推理模型以及闭源非推理API，表示了部分baseline模型。

\paragraph{\textbf{LLMs tend to adopt explicit reasoning paradigms when performing analysis and problem-solving}}

Response-level case studies reveal that LLMs in KORGym employ four primary reasoning paradigms, each reflecting a distinct cognitive strategy:
\begin{itemize}[leftmargin=6mm]
    \item \textbf{Code Paradigm}: generating executable code to obtain solutions (e.g., ``import math; a = 0; for i in range(...): ...'').
    \item \textbf{Mathematical Paradigm}: applying algebraic equations or arithmetic rules to model and solve problems (e.g., ``Let x be the number of creature A and y the number of creature B; construct the system of equations…'').
    \item \textbf{Algorithm-Specific Paradigm}: invoking established algorithms (e.g., Dijkstra’s algorithm, Eulerian path) and adapting them to the task context (e.g., ``Use an Eulerian path to solve the one-stroke drawing puzzle: first compute …'').
    \item \textbf{Natural Language Reasoning Paradigm}: conducting spatial, logical, or causal analysis in natural language (e.g., ``If we turn right, we reach (1,2) where a springboard lies ahead …'').
\end{itemize}

\begin{figure*}[ht]
  % 可选参数 [n] 指定环绕图片的文本行数，比如 [14]
  % \begin{wrapfigure}[14]{r}{0.5\textwidth}

  \centering
  \includegraphics[width=0.9\linewidth]{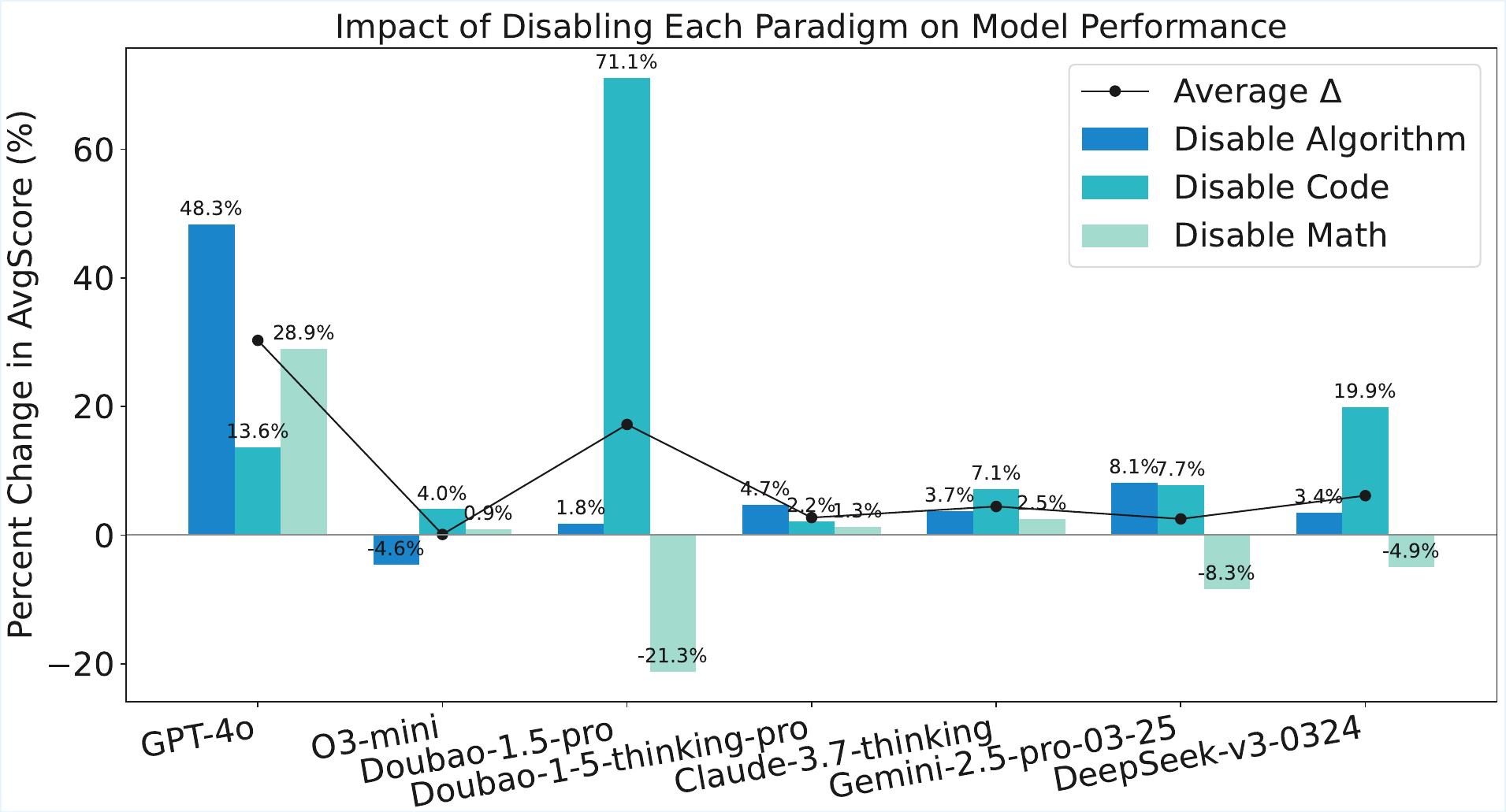}
  \caption{Impact of Disabling Each Paradigm on Model Performance}
  \label{fig:paradigm}
\end{figure*}

%为了进一步对Reasoning Paradigm进行分析，我们选用 GPT-4o作为LLM Judger，令其对LLM在KORGYM中部分游戏对应的responses进行分类，并统计每种回答的平均分数，由此得到如表所示的数据：
To examine reasoning-paradigm usage, we employ GPT-4o to annotate model responses for selected KORGym games and compute the mean score for each paradigm (Table~\ref{tab:paradigm}). We then conduct ablation experiments by constraining prompts to disable individual paradigms, with results summarized in Figure~\ref{fig:paradigm}. Our key findings include:

\input{tables/analysis/paradigm}

% \begin{figure*}[ht]
% \centering
% \includegraphics[width=0.9\textwidth]{pics/analysis/paradigm.pdf}
% \caption{Impact of Disabling Each Paradigm on Model Performance}
% \label{fig:paradigm}
% \end{figure*}

%我们观察到如下的现象：-每个series的模型在推理过程中展现出其行为模式的偏好 在推理过程中，Gemini-2.5-pro倾向于使用Code作为解题范式，而O3-mini则倾向于使用math与自然语言推理作为解题范式，Doubao-1-5-thinking-pro 倾向于利用Algorithm进行解题 -推理范式在一定程度上限制了模型推理能力 在禁用推理范式后，所有模型的平均得分均有所上升，我们推测这是由于大量math/code/algorithm相关的数据集被用于模型预训练以提高模型分数导致的模型针对全新场景的分析及推理能力下降，同时，也侧面证明了我们KORGYM提供了一个全新的检验模型推理能力的场景。-数学推理为模型推理过程的重要组成 禁用数学推理范式后，大多数模型得分下降/无明显变化，这说明数学推理为模型推理过程中最为重要的组成部分。 -较强模型往往具备较强的鲁棒性 能力越强的模型受禁用推理范式的影响越小，如o3-mini,这说明较强的模型拥有更强的鲁棒性与泛化性，且能根据具体情况对问题进行分析与推理

\paragraph{\textbf{Models within the same series exhibit distinct reasoning-paradigm preferences}}
Models tend to adopt paradigms aligned with their architecture: Gemini-2.5-Pro predominantly employs code-based reasoning; O3-mini primarily utilizes mathematical and natural language reasoning; and Doubao-1.5-thinking-pro strongly favors algorithm-specific reasoning strategies.

\paragraph{\textbf{Reasoning Paradigms Partially Constrain Model Performance}}
Disabling specific reasoning paradigms via prompt constraints led to increased average performance across all models. We attribute this improvement to overreliance on large-scale pretraining data rich in mathematical, coding, and algorithmic examples, which can impede generalization and adaptability to novel reasoning tasks. These findings underscore KORGym’s value as a robust benchmark for evaluating genuine reasoning abilities beyond memorized patterns.

\paragraph{\textbf{Mathematical Reasoning as a Core Component of the Reasoning Process}}
Disabling the mathematical paradigm causes most models to experience a performance decline or exhibit no improvement, indicating that mathematical reasoning is critical to LLM reasoning capabilities.

\paragraph{\textbf{Stronger Models Exhibit Greater Robustness}}
More capable models are less impacted by disabling individual reasoning paradigms. For example, O3-mini and Doubao-1.5-thinking-pro maintain near-original performance when deprived of their preferred reasoning strategies, demonstrating superior robustness, generalization, and adaptive reasoning under constraint.

\subsection{RQ3: What is the Impact of Reinforcement Learning (RL) on Problem Solving Capabilities?}
During multi-turn reinforcement-learning fine-tuning, Doubao-1.5-thinking-pro incorporated two specialized algorithmic frameworks—DAPO and VAPO—to address instability in reasoning-oriented model training. In parallel, it was trained on a comprehensive corpus of \textbf{STEM problems, code-related tasks, logical reasoning challenges, and non-reasoning examples}. Additionally, RL training on classic games (e.g., 24-point, mazes, and Sudoku) yielded a marked improvement in its reasoning performance.

\begin{wrapfigure}[15]{r}{0.6\textwidth}
  \centering
  \vspace{-5mm}
  \includegraphics[width=0.85\linewidth]{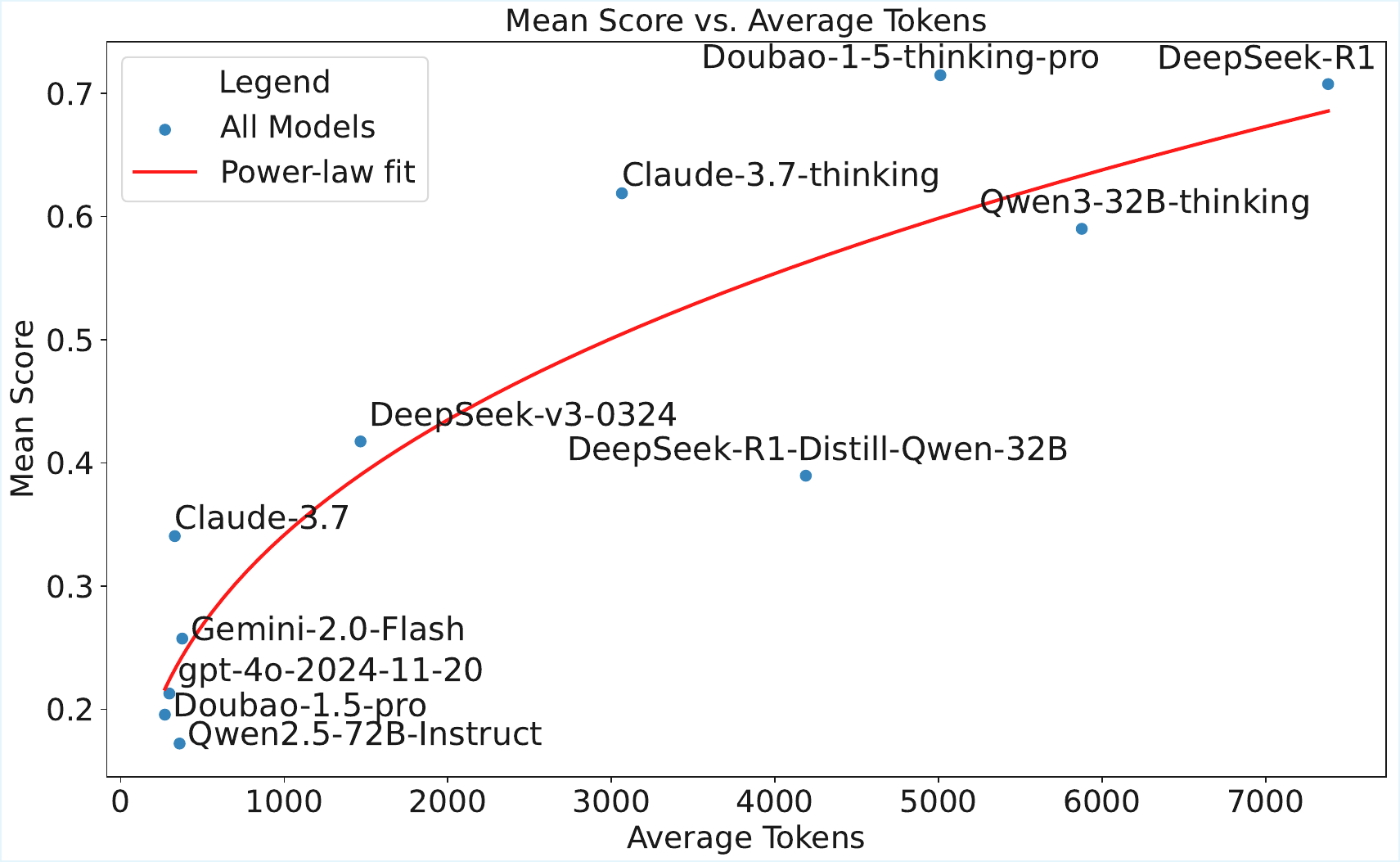}
  \caption{Correlation Between Response Length and Reasoning Performance}
  \label{fig:token_analysis}
\end{wrapfigure}

In KORGym, RL-driven enhancements yielded substantial gains across reasoning dimensions. As shown in Table~\ref{tab:overall_performance}, Doubao-1.5-thinking-pro achieved a mean score of 0.72—fourth overall—and excelled in puzzle reasoning with a score of 0.84, surpassing both O1 and O3-mini. Notably, Doubao-1.5-thinking-pro exhibits minimal performance degradation under ablation and demonstrates score variance comparable to leading models (e.g., O3-mini and O1-2024-12-17). These improvements underscore that appropriate reinforcement learning fosters both enhanced reasoning and more balanced performance.

% More importantly, Doubao-1.5-thinking-pro exhibited minimal performance degradation in the reasoning paradigm ablation experiments, and its score variance across different game scenarios is on par with top-tier RL-optimized models such as O3-mini and O1-2024-12-17.
% This demonstrates that \textbf{through hybrid-paradigm RL training, the model not only excels in individual reasoning skills but also maintains balanced and stable performance across diverse problem-solving strategies}.

%Doubao-1-5-thinking-pro 在多轮次强化学习（RL）微调阶段，通过引入两种专门针对推理模型训练不稳定问题的算法框架——DAPO（policy-gradient）与 VAPO（actor-critic），并由此设计了完善的STEM问题、代码相关任务、逻辑推理以及非推理数据的数据集，并对常见的games,如24-point, mazes, Sudoku进行了RL训练，使模型能力出现了极大的飞跃。
%而在 KORGym 实验中，这些 RL 驱动的改进带来了跨维度的 推理能力飞跃：如Table 4所示，Doubao-1-5-thinking-pro的Average Score达到0.72,位列第四，而Puzzle Reasoning更是达到0.840，超越了O1和O3-mini。这些巨大增益直接反映了策略优化对长链式、多步推理能力的有效促进。
%更为关键的是，Doubao-1-5-thinking-pro 在各推理范式消融实验中展现了极小的性能波动，同时其在不同游戏场景下的得分方差水平已与 O3-mini、O1-2024-12-17 等顶级 RL 模型持平，表明通过混合范式的 RL 训练，它不仅在单一能力上表现卓越，更在多种解题策略间维持了均衡、稳定的推理能力。由此可见，RL对Problem-Solving Capabilities的能力与鲁棒性均有极强的促进作用。

\subsection{RQ4: Is There a Correlation Between Response Length and Reasoning Performance?}

%为了探究Response Length 对 Reasoning Performance的影响，我们统计了4个推理模型和8个非推理模型在进行game推理过程中的token长度及得分，并由此进行曲线拟合，如图所示。
To examine the effect of response length on reasoning performance, we record token counts and reasoning scores for four reasoning models and eight non-reasoning models during gameplay. We then fit curves to the aggregated data to identify trends and correlations, as illustrated in Figure~\ref{fig:token_analysis}.

%根据该图像我们可以观察到：- Reasoning Performance和Response Length具有很明显的相关性，且token长度越长，模型在KORGym得分越高；- 推理模型和非推理模型存在极大的Response Length长度差距，且非推理模型response长度较为集中，推理模型response长度较为分散；-Response Length 对 Reasoning Performance的提升存在边际效应 随着response长度的进一步上升，模型平均得分的上升逐渐趋于平缓。

From the figure, we derive the following insights:
\begin{itemize}[leftmargin=6mm]
    \item \textbf{A strong positive correlation exists between reasoning performance and response length}: models with longer responses tend to achieve higher KORGym scores.
    \item \textbf{Reasoning and non-reasoning models differ markedly in response length distributions}: non-reasoning models produce responses within a narrow range, whereas reasoning models exhibit a broader and more varied distribution.
    \item \textbf{The impact of response length on performance exhibits diminishing returns}: as response length increases, incremental score gains become marginal, suggesting an upper limit to the benefits of verbosity.
\end{itemize}

%% file: tables/analysis/paradigm.tex
% 需要在导言区引入：
% \usepackage{booktabs}
% \usepackage{multirow}

\begin{table}[h]
    \centering
    \small
    \caption{ Reasoning Paradigm Proportions and Average Scores for Different Models' Responses }
    \resizebox{0.99\textwidth}{!}{
    \begin{tabular}{lccccccccc}
        \toprule
        % 第一行
        \multirow{2}{*}{Model}
        & \multicolumn{4}{c}{Proportion (\%)}
        & \multicolumn{4}{c}{Average Score(\%)}
        & \multirow{2}{*}{Overall Score(\%)} \\
        \cmidrule(lr){2-5} \cmidrule(lr){6-9}
        % 第二行
        & Code 
        & Math 
        & Algorithm 
        & Natural Language 
        & Code 
        & Mathematical 
        & Algorithm 
        & Natural Language 
        &  \\  % 占位
        \midrule
        Doubao-1-5-thinking-pro&	0.6&	14&	32.9	&52.6&	50	&82	&0.61	&66&	66.565\\
Doubao-1.5-pro& 	17.1&	10.3	&42.6&	30&	0&	0	&6&	6	&4.356\\
DeepSeek-v3-0324&	0	&0&	9.7&	90.3&	0&	0	&26&	22&	22.388\\
Claude-3.7-thinking&	0.3&	13.4&	26.3&	60&	0	&64&	37&	46&	45.907\\
Gemini-2.5-pro-03-25&	46&	6.9&	12&	35.1&	50	&83&	100&	66&	63.893\\
GPT-4o&	2&	1.1&	29.7&	67.1&	0&	0&	2&	2&	1.936\\
O3-mini&	0&	21.4&	11.4&	67.1&	0&	85&	97&	72&	77.56\\

        \bottomrule
    \end{tabular}
    }

    \label{tab:paradigm}
\end{table}

%% file: sections/conclusion.tex
\section{Conclusion}
%在本文，我们提出了KORGym-A scalable, game-driven benchmark with 50+ tasks across six reasoning dimensions.同时，我们的benchmark支持multimodal interactions, reinforcement learning, and parameterized environments，并基于游戏特点提出了A robust evaluation methodology with dimension-aware score aggregation.在实验部分，我们利用19LLMs和5VLMs进行了评估实验，发现Similar Strength–Weakness Profiles Within Same Model Series以及Model Scale and Architecture on Reasoning Capabilities。此后，我们对 Modality，Model Series，Reinforcement Learning 以及einforcement Learning 对Reasoning Performance的影响展开了消融实验，对模型easoning Capabilities的影响因素进行了详细的分析与讨论。在未来，我们将进一步扩充KORGym的游戏数量与广度，并将引入对抗性游戏以及合作性游戏以对模型推理能力进行更深层次的评测与分析。希望我们的工作对推动LLM推理能力以及RL工作起到积极促进作用。

In this paper, we introduce KORGym—a scalable, game-driven benchmark comprising over fifty tasks spanning six reasoning dimensions. KORGym supports multimodal interactions, reinforcement learning, and parameterized environments, and employs a robust evaluation methodology based on dimension-aware score aggregation tailored to game-based reasoning. We evaluate 19 LLMs and 8 VLMs, revealing consistent strength–weakness profiles within model series and demonstrating the impact of model scale and architecture on reasoning capabilities. We also conduct ablation studies on modality, model series, and reinforcement learning, providing a detailed analysis of the key factors influencing LLM reasoning performance.

% Due to time constraints, KORGym’s current selection of games may lack quantity and diversity, representing a potential limitation. To address this, we will keep expanding KORGym by adding more varied games, thereby enabling deeper evaluation and more interactive reasoning capabilities. Looking ahead, we plan to expand KORGym by increasing both the number and diversity of games, and by incorporating adversarial and collaborative gameplay to enable deeper evaluation of multi-agent and interactive reasoning capabilities.
% We hope our work contributes to advancing both LLM reasoning and RL-based evaluation in open-ended, dynamic environments.

%% file: sections/appendix.tex
\section{Contributions and Acknowledgements}
\textbf{Leading Authors}

\begin{itemize}
    \item Jiajun Shi, ByteDance Seed, M-A-P, Beihang University
    \item Jian Yang, M-A-P, Beihang University
    \item Jiaheng Liu, M-A-P
    \item Ge Zhang, ByteDance Seed, M-A-P
\end{itemize}

\textbf{Corresponding Authors}
\begin{itemize}
    \item Wenhao Huang, ByteDance Seed
    \item Ge Zhang, ByteDance Seed, M-A-P
\end{itemize}

\textbf{Contributors}
\begin{itemize}
    \item Xingyuan Bu, M-A-P
    \item Jiangjie Chen, ByteDance Seed
    \item Junting Zhou, M-A-P
    \item Kaijing Ma, ByteDance Seed, M-A-P
    \item Zhoufutu Wen, ByteDance Seed, M-A-P
    \item Bingli Wang, M-A-P
    \item Yancheng He, M-A-P
    \item Liang Song, M-A-P
    \item Hualei Zhu, M-A-P
    \item Shilong Li, M-A-P
    \item Xingjian Wang, M-A-P
    \item Wei Zhang, M-A-P
    \item Ruibin Yuan, M-A-P
    \item Yifan Yao, M-A-P
    \item Wenjun Yang, M-A-P
    \item Yunli Wang, M-A-P
    \item Siyuan Fang, M-A-P
    \item Siyu Yuan, ByteDance Seed
    \item Qianyu He, ByteDance Seed
    \item Xiangru Tang, M-A-P
    \item Yingshui Tan, M-A-P
    \item Wangchunshu Zhou
    \item Zhaoxiang Zhang
    \item Zhoujun Li, Beihang University
\end{itemize}

\textbf{Special Acknowledgements}
We sincerely thank Minchao Wang, Zaiyuan Wang, and Liang Hu for their invaluable support and insightful guidance on the development of KORGym. We also extend our gratitude to ByteDance and all members of MAP who have contributed their expertise and effort toward the success and advancement of the KORGym project. Their dedication, collaboration, and valuable suggestions have been instrumental in shaping this research and enabling its impactful outcomes.

\section{Formal Definition of ``Knowledge Orthogonality''}\label{kor}

\textbf{For a task $T$, the required reasoning information consists of:}
\begin{itemize}
    \item $K$: General background/domain-specific knowledge acquired during pre-training, excluding common sense.
    \item $R$: Core rule information designed to solve $T$.
    \item $Q$: A Rule-Driven question.
    \item $A$: Answer to the question $Q$.
\end{itemize}

\textbf{Notational Definitions:}
\begin{itemize}
    \item $\rightarrow$: Represents the cognitive process of deriving $A$ from $Q$.
    \item $P$: Represents the belief strength that $A$ is a valid answer to $Q$ based on $R$ and/or $K$.
    \begin{itemize}
        \item $P(Q \rightarrow A \mid R)$: Belief in $A$ driven solely by $R$.  
        \item $P(Q \rightarrow A \mid K)$: Belief in $A$ based solely on $K$.  
        \item $P(Q \rightarrow A \mid R, K)$: Combined belief in $A$, integrating $R$ and $K$.
    \end{itemize}
\end{itemize}

\textbf{$T$ satisfies knowledge orthogonality under the following conditions:}
\begin{enumerate}
    \item \textbf{Knowledge-Rule Decoupling}: Rule $R$ is logically self-contained and independent of $K$.
    \[
    R \perp K
    \]
    \item \textbf{Knowledge Assistiveness}: Background knowledge $K$ may support or interfere with the derivation of $A$ from $Q$, but does not play a central role in reasoning. The extent of this influence is quantified by the Knowledge Impact Factor ($\beta$), defined as:
    \[
    \beta = \frac{P(Q \rightarrow A \mid R, K) - P(Q \rightarrow A \mid R)}{P(Q \rightarrow A \mid R)}
    \]
    $\beta$ ranges from $(-1, \epsilon]$, where $\epsilon$ is a very small positive number.
    \begin{itemize}
        \item When $\beta$ is positive and close to 0, $K$ has little impact, with $R$ being dominant.
        \item When $\beta$ is negative, it can range from small negative values to approaching $-1$, where $K$ increasingly undermines reasoning.
    \end{itemize}
    \item \textbf{Rule Centrality}: Correctness relies on understanding and applying $R$, with $R$ having significantly greater influence than $K$.
    \[
    P(Q \rightarrow A \mid R, K) \approx P(Q \rightarrow A \mid R) \gg P(Q \rightarrow A \mid K)
    \]
    \item \textbf{Derivation Adjustment}: This formula adjusts the reasoning process based on $R$, incorporating the influence of $K$ with $\beta$ reflecting its effect.
    \[
    P(Q \rightarrow A \mid R, K) = P(Q \rightarrow A \mid R) \cdot (1 + \beta)
    \]
\end{enumerate}

\newpage
\section{Detailed Scores on KORGym}\label{Appendix:detailed_scores}
%\subsection{Hello World}

\input{tables/exp/math}
\input{tables/exp/control}
\input{tables/exp/language}
\input{tables/exp/spatial}
\input{tables/exp/strategy}

\newpage

\section{PCA Result}
The first two principal components capture 96.2\% of the total variance (PC1: 91.9\%, PC2: 4.3\%), as shown in Figure \ref{fig:PCA}. The model PCA cluster results are as follows:
\begin{itemize}
    \item \textbf{Cluster 0}:Doubao-1-5-thinking-pro, DeepSeek-R1, Claude-3.7-thinking, Gemini-2.5-pro-03-25, Qwen3-32B
    \item \textbf{Cluster 1}:O3-mini, O1-2024-12-17
    \item \textbf{Cluster 2}:Doubao-1.5-pro, DeepSeek-R1-Distill-Qwen-7B, GPT-4o, Gemini-2.0-Flash, Qwen2.5-72B-Instruct, Qwen2.5-32B-Instruct, Qwen2.5-7B-Instruct
    \item \textbf{Cluster 3}:: DeepSeek-R1-Distill-Qwen-32B, DeepSeek-v3-0324, Claude-3.7, Gemini-2.0-Flash-thinking, Qwen-QwQ
\end{itemize}

These results reveal the following insights from the clustering structure:
\begin{itemize}
    \item \textbf{Cluster 1 predominantly consists of the GPT-o series reasoning models}, which exhibit highly similar reasoning patterns, indicating a consistent architectural behavior within the series.
    \item \textbf{Cluster 0 includes top-performing reasoning models}, such as Claude-3.7, Gemini-2.5, and DeepSeek-R1, which show strong and mutually consistent performance across dimensions.
    \item \textbf{Qwen3 and Doubao-1.5-thinking-pro are located near the boundary between Cluster 0 and Cluster 1}, suggesting that these models share reasoning characteristics with both the GPT-o series and the leading reasoning models in Cluster 0.
    \item \textbf{Cluster 3 primarily contains a mix of open-source reasoning models and closed-source non-reasoning models}, exhibiting moderate overall reasoning ability.
    \item \textbf{Cluster 2 consists of mainstream open-source non-reasoning models and closed-source baseline APIs}, representing a group of lower-performing or general-purpose models.
\end{itemize}

%% file: tables/exp/math.tex
\begin{table}[h]
    \centering
    \small
    \caption{Mathematical and logical reasoning abilities  of different models on KORGym.}
    \resizebox{0.99\textwidth}{!}{
    \begin{tabular}{lccccccccccc}
        \toprule
        Model & \makecell{Date\\Calculation} & Sudoku & \makecell{Light\\Out Game} & \makecell{Square\\Addition} & Alien & \makecell{Party\\Time} & \makecell{Path Planning\\Problem} & \makecell{Construction\\Company} & \makecell{One\\Stroke Drawing} & \makecell{Coloring\\Issue} & \makecell{City\\Traveller} \\
        \midrule
Doubao-1.5-pro                   & 0.060 & 0.120 & 0.060 & 0.000 & 0.200 & 0.240 & 0.140 & 0.020 & 0.060 & 0.060 & 0.340 \\
Doubao-1-5-thinking-pro  & 0.160 & 0.600 & 0.860 & 0.800 & 0.880 & 0.640 & 0.800 & 0.920 & 0.340 & 0.980 & 0.680 \\
DeepSeek-R1-Distill-Qwen-32B    & 0.120 & 0.200 & 0.260 & 0.220 & 0.400 & 0.540 & 0.120 & 0.100 & 0.020 & 0.700 & 0.380 \\
DeepSeek-R1-Distill-Qwen-7B     & 0.060 & 0.000 & 0.060 & 0.000 & 0.120 & 0.020 & 0.020 & 0.000 & 0.000 & 0.140 & 0.000 \\
DeepSeek-R1                  & 0.200 & 0.420 & 0.540 & 0.680 & 0.740 & 0.520 & 0.620 & 0.700 & 0.380 & 0.940 & 0.460 \\
DeepSeek-v3-0324             & 0.120 & 0.020 & 0.300 & 0.460 & 0.260 & 0.140 & 0.200 & 0.160 & 0.200 & 0.140 & 0.320 \\
Claude-3.7               & 0.040 & 0.140 & 0.060 & 0.060 & 0.340 & 0.280 & 0.180 & 0.120 & 0.180 & 0.200 & 0.700 \\
Claude-3.7-thinking      & 0.080 & 0.100 & 0.160 & 0.640 & 0.720 & 0.880 & 0.480 & 0.500 & 0.680 & 0.700 & 0.000 \\
GPT-4o                   & 0.000 & 0.060 & 0.040 & 0.000 & 0.140 & 0.080 & 0.020 & 0.020 & 0.020 & 0.020 & 0.280 \\
O3-mini                  & 0.160 & 0.420 & 0.980 & 0.880 & 0.720 & 0.780 & 0.600 & 0.860 & 0.760 & 0.960 & 0.280 \\
O1-2024-12-17            & 0.260 & 0.040 & 1.000 & 0.700 & 0.840 & 0.600 & 0.340 & 0.280 & 0.760 & 0.800 & 0.300 \\
Gemini-2.0-Flash          & 0.020 & 0.160 & 0.040 & 0.060 & 0.160 & 0.280 & 0.200 & 0.000 & 0.000 & 0.080 & 0.440 \\
Gemini-2.5-pro-03-25     & 0.240 & 0.640 & 0.800 & 0.760 & 0.780 & 0.920 & 0.540 & 0.920 & 1.000 & 0.980 & 0.840 \\
Gemini-2.0-Flash-thinking & 0.140 & 0.100 & 0.120 & 0.140 & 0.220 & 0.560 & 0.140 & 0.040 & 0.320 & 0.740 & 0.500 \\
Qwen-QwQ                 & 0.080 & 0.080 & 0.080 & 0.300 & 0.240 & 0.000 & 0.020 & 0.000 & 0.000 & 0.340 & 0.000 \\
Qwen2.5-72B-Instruct     & 0.000 & 0.020 & 0.060 & 0.000 & 0.120 & 0.000 & 0.060 & 0.060 & 0.020 & 0.040 & 0.000 \\
Qwen2.5-32B-Instruct     & 0.000 & 0.040 & 0.060 & 0.000 & 0.120 & 0.060 & 0.060 & 0.000 & 0.000 & 0.020 & 0.000 \\
Qwen2.5-7B-Instruct      & 0.000 & 0.000 & 0.000 & 0.000 & 0.000 & 0.000 & 0.000 & 0.000 & 0.000 & 0.000 & 0.020 \\
        \bottomrule
    \end{tabular}
    }
    
    \label{tab:math_game_transposed}
\end{table}

%% file: tables/exp/control.tex
\begin{table}[h]
    \centering
    \small
    \caption{Control, interaction, and task reasoning abilities  of different models on KORGym.}
    \resizebox{0.99\textwidth}{!}{
    \begin{tabular}{lcccccccccc}
        \toprule
        Model & Minigrid & Snake & Tetris & \makecell{Tower\\of Hanoi} & \makecell{Numeric\\Bricks} & Minesweeper & Nullify & PVZ & Long Cat & \makecell{Black\\White Copy} \\
        \midrule
Doubao-1.5-pro                    & 0.050 & 1.800 & 0.000 & 0.150 & 0.000 & 0.000 & 0.250 & 17.550 & 0.020 & 0.000 \\
Doubao-1-5-thinking-pro  & 0.300 & 16.300 & 0.300 & 0.650 & 0.000 & 0.076 & 0.400 & 59.950 & 0.600 & 0.360 \\
DeepSeek-R1-Distill-Qwen-32B    & 0.100 & 7.480 & 0.300 & 0.700 & 0.000 & 0.162 & 0.300 & 31.750 & 0.220 & 0.200 \\
DeepSeek-R1-Distill-Qwen-7B     & 0.120 & 0.150 & 0.000 & 0.050 & 0.000 & 0.042 & 0.100 & 11.950 & 0.000 & 0.000 \\
DeepSeek-R1                  & 0.100 & 16.500 & 0.650 & 0.700 & 0.020 & 0.185 & 0.200 & 68.450 & 0.480 & 0.300 \\
DeepSeek-v3-0324             & 0.250 & 4.250 & 0.050 & 0.200 & 0.000 & 0.063 & 0.300 & 20.350 & 0.300 & 0.080 \\
Claude-3.7               & 0.150 & 6.900 & 0.100 & 0.100 & 0.000 & 0.049 & 0.150 & 27.300 & 0.140 & 0.020 \\
Claude-3.7-thinking      & 0.350 & 9.750 & 0.150 & 0.500 & 0.000 & 0.297 & 0.250 & 40.300 & 0.320 & 0.180 \\
GPT-4o                   & 0.000 & 0.600 & 0.000 & 0.050 & 0.000 & 0.027 & 0.150 & 25.000 & 0.000 & 0.000 \\
O3-mini                  & 0.300 & 15.950 & 0.900 & 1.000 & 0.000 & 0.555 & 0.250 & 41.200 & 0.700 & 0.360 \\
O1-2024-12-17            & 0.150 & 17.500 & 0.550 & 0.850 & 0.000 & 0.741 & 0.300 & 54.850 & 0.840 & 0.360 \\
Gemini-2.0-Flash          & 0.150 & 2.050 & 0.000 & 0.100 & 0.000 & 0.008 & 0.300 & 19.650 & 0.100 & 0.020 \\
Gemini-2.5-pro-03-25     & 0.400 & 13.500 & 0.050 & 0.650 & 0.000 & 0.235 & 0.300 & 61.300 & 0.600 & 0.380 \\
Gemini-2.0-Flash-thinking & 0.100 & 1.850 & 0.050 & 0.200 & 0.000 & 0.002 & 0.250 & 26.000 & 0.080 & 0.100 \\
Qwen-QwQ                 & 0.150 & 3.950 & 0.000 & 0.600 & 0.000 & 0.052 & 0.350 & 30.750 & 0.020 & 0.160 \\
Qwen2.5-72B-Instruct     & 0.050 & 0.620 & 0.000 & 0.200 & 0.000 & 0.000 & 0.250 & 24.850 & 0.040 & 0.000 \\
Qwen2.5-32B-Instruct     & 0.050 & 0.300 & 0.000 & 0.050 & 0.000 & 0.021 & 0.200 & 14.450 & 0.020 & 0.020 \\
Qwen2.5-7B-Instruct      & 0.000 & 0.450 & 0.000 & 0.050 & 0.000 & 0.000 & 0.250 & 0.050 & 0.000 & 0.020 \\
        \bottomrule
    \end{tabular}
    }
    
    \label{tab:control_game_transposed}
\end{table}

%% file: tables/exp/language.tex
\begin{table}[h]
    \centering
    \small
    \caption{Language and textual reasoning abilities  of different models on KORGym.}
    \resizebox{0.99\textwidth}{!}{
    \begin{tabular}{lccccccc}
        \toprule
        Model & Word Problem & Alphabetical sorting & Letter Connection & Word Transformation & Wordle & Crypto Word \\
        \midrule
Doubao-1.5-pro                    & 0.120 & 0.360 & 0.160 & 0.140 & 0.100 & 0.150 \\
Doubao-1-5-thinking-pro  & 0.600 & 0.880 & 0.720 & 0.400 & 0.600 & 1.000 \\
DeepSeek-R1-Distill-Qwen-32B    & 0.340 & 0.500 & 0.220 & 0.140 & 0.450 & 0.000 \\
DeepSeek-R1-Distill-Qwen-7B     & 0.020 & 0.240 & 0.020 & 0.000 & 0.050 & 0.000 \\
DeepSeek-R1                  & 0.820 & 0.960 & 0.900 & 0.420 & 0.600 & 0.950 \\
DeepSeek-v3-0324             & 0.460 & 0.840 & 0.420 & 0.380 & 0.500 & 0.500 \\
Claude-3.7               & 0.580 & 0.840 & 0.660 & 0.220 & 0.250 & 0.650 \\
Claude-3.7-thinking      & 0.820 & 0.980 & 0.960 & 0.560 & 0.850 & 1.000 \\
GPT-4o                   & 0.420 & 0.340 & 0.160 & 0.120 & 0.400 & 0.100 \\
O3-mini                  & 0.880 & 0.980 & 0.980 & 0.400 & 0.400 & 1.000 \\
O1-2024-12-17            & 0.960 & 1.000 & 0.980 & 0.480 & 0.450 & 0.850 \\
Gemini-2.0-Flash          & 0.340 & 0.560 & 0.340 & 0.120 & 0.250 & 0.100 \\
Gemini-2.5-pro-03-25     & 0.900 & 0.960 & 0.940 & 0.780 & 0.650 & 1.000 \\
Gemini-2.0-Flash-thinking & 0.620 & 0.780 & 0.400 & 0.180 & 0.550 & 0.500 \\
Qwen-QwQ                 & 0.480 & 0.760 & 0.400 & 0.180 & 0.400 & 0.050 \\
Qwen2.5-72B-Instruct     & 0.080 & 0.280 & 0.160 & 0.020 & 0.200 & 0.000 \\
Qwen2.5-32B-Instruct     & 0.100 & 0.280 & 0.080 & 0.000 & 0.050 & 0.050 \\
Qwen2.5-7B-Instruct      & 0.040 & 0.140 & 0.020 & 0.000 & 0.050 & 0.000 \\
        \bottomrule
    \end{tabular}
    }
    
    \label{tab:language_game_transposed}
\end{table}

%% file: tables/exp/spatial.tex
\begin{table}[h]
    \centering
    \small
    \caption{Spatial and geometric reasoning abilities  of different models on KORGym.}
    \resizebox{0.99\textwidth}{!}{
    \begin{tabular}{lccccccccccc}
        \toprule
        Model & Maze & Sokoban & play lines & \makecell{Emoji\\Connect} & 8-puzzle & \makecell{Bubble Ball\\Sorting} & \makecell{Pipe\\Game} & \makecell{Free\\the Key} & \makecell{Map\\Simulation} & \makecell{Arrow Pathway} \\
        \midrule
Doubao-1.5-pro                    & 0.000 & 0.000 & 0.080 & 0.080 & 0.080 & 0.300 & 0.000 & 0.050 & 0.020 & 0.000 \\
Doubao-1-5-thinking-pro  & 0.800 & 0.420 & 0.580 & 0.580 & 0.680 & 1.000 & 0.700 & 0.950 & 0.320 & 0.000 \\
DeepSeek-R1-Distill-Qwen-32B    & 0.380 & 0.160 & 0.140 & 0.220 & 0.380 & 0.900 & 0.040 & 0.700 & 0.040 & 0.000 \\
DeepSeek-R1-Distill-Qwen-7B     & 0.120 & 0.040 & 0.000 & 0.000 & 0.100 & 0.000 & 0.000 & 0.000 & 0.000 & 0.000 \\
DeepSeek-R1                  & 0.600 & 0.440 & 0.360 & 0.760 & 0.520 & 1.000 & 0.300 & 0.750 & 0.140 & 0.000 \\
DeepSeek-v3-0324             & 0.380 & 0.080 & 0.080 & 0.560 & 0.220 & 0.250 & 0.020 & 0.650 & 0.060 & 0.000 \\
Claude-3.7               & 0.020 & 0.020 & 0.020 & 0.160 & 0.140 & 0.850 & 0.000 & 0.200 & 0.040 & 0.020 \\
Claude-3.7-thinking      & 0.400 & 0.260 & 0.240 & 0.760 & 0.460 & 0.950 & 0.100 & 0.850 & 0.120 & 0.180 \\
GPT-4o                   & 0.020 & 0.040 & 0.020 & 0.120 & 0.040 & 0.450 & 0.000 & 0.300 & 0.000 & 0.000 \\
O3-mini                  & 0.860 & 0.780 & 0.720 & 0.960 & 0.940 & 0.950 & 0.540 & 0.850 & 0.380 & 0.300 \\
O1-2024-12-17            & 0.360 & 0.700 & 0.840 & 0.940 & 0.900 & 0.950 & 0.840 & 0.850 & 0.280 & 0.000 \\
Gemini-2.0-Flash          & 0.020 & 0.000 & 0.000 & 0.120 & 0.040 & 0.600 & 0.000 & 0.050 & 0.100 & 0.000 \\
Gemini-2.5-pro-03-25     & 0.500 & 0.100 & 0.660 & 0.620 & 0.720 & 0.950 & 0.740 & 0.850 & 0.100 & 0.000 \\
Gemini-2.0-Flash-thinking & 0.020 & 0.060 & 0.000 & 0.180 & 0.480 & 0.650 & 0.000 & 0.200 & 0.020 & 0.020 \\
Qwen-QwQ                 & 0.400 & 0.180 & 0.080 & 0.060 & 0.240 & 0.000 & 0.000 & 0.650 & 0.000 & 0.000 \\
Qwen2.5-72B-Instruct     & 0.040 & 0.020 & 0.020 & 0.060 & 0.040 & 0.350 & 0.000 & 0.100 & 0.000 & 0.000 \\
Qwen2.5-32B-Instruct     & 0.020 & 0.000 & 0.040 & 0.000 & 0.000 & 0.500 & 0.000 & 0.200 & 0.020 & 0.000 \\
Qwen2.5-7B-Instruct      & 0.000 & 0.000 & 0.000 & 0.020 & 0.000 & 0.000 & 0.000 & 0.000 & 0.000 & 0.000 \\
        \bottomrule
    \end{tabular}
    }
    
    \label{tab:spatial_game_transposed}
\end{table}

%% file: tables/exp/strategy.tex
\begin{table}[h]
    \centering
    \small
    \caption{Strategic and game-theoretic reasoning abilities of different models on KORGym.}
    \resizebox{0.99\textwidth}{!}{
    \begin{tabular}{lccccccccccc}
        \toprule
        Model  & 2048 & Trust Evolution & N point & Spider Solitaire & Circle the cat \\
        \midrule
        
Doubao-1.5-pro                    & 648.800 & 31.650 & 7.500 & 0.000 & 0.000 \\
Doubao-1-5-thinking-pro  & 1736.000 & 36.600 & 8.000 & 0.000 & 0.050 \\
DeepSeek-R1-Distill-Qwen-32B    & 990.000 & 37.700 & 8.050 & 0.000 & 0.050 \\
DeepSeek-R1-Distill-Qwen-7B     & 884.200 & 36.350 & 3.625 & 0.000 & 0.000 \\
DeepSeek-R1                  & 2221.800 & 63.500 & 8.250 & 0.000 & 0.200 \\
DeepSeek-v3-0324             & 1101.400 & 37.750 & 8.225 & 0.000 & 0.200 \\
Claude-3.7               & 809.000 & 36.900 & 8.300 & 0.000 & 0.000 \\
Claude-3.7-thinking      & 1283.600 & 37.950 & 8.200 & 0.000 & 0.100 \\
GPT-4o                   & 958.600 & 54.750 & 7.650 & 0.000 & 0.000 \\
O3-mini                  & 1285.600 & 34.550 & 7.825 & 0.000 & 0.300 \\
O1-2024-12-17            & 1652.200 & 64.500 & 8.600 & 0.000 & 0.000 \\
Gemini-2.0-Flash          & 879.800 & 51.750 & 7.475 & 0.000 & 0.000 \\
Gemini-2.5-pro-03-25     & 2083.200 & 66.500 & 7.750 & 0.000 & 0.250 \\
Gemini-2.0-Flash-thinking & 1378.200 & 37.500 & 7.925 & 0.000 & 0.000 \\
Qwen-QwQ                 & 681.800 & 58.400 & 3.950 & 0.000 & 0.000 \\
Qwen2.5-72B-Instruct     & 773.400 & 47.150 & 7.600 & 0.000 & 0.000 \\
Qwen2.5-32B-Instruct     & 660.000 & 51.850 & 7.550 & 0.000 & 0.000 \\
Qwen2.5-7B-Instruct      & 780.600 & 0.000 & 6.200 & 0.000 & 0.050 \\
        \bottomrule
    \end{tabular}
    }
    
    \label{tab:strategy_game_transposed}
\end{table}

%% file: paper.bbl
\begin{thebibliography}{43}
\providecommand{\natexlab}[1]{#1}
\providecommand{\url}[1]{\texttt{#1}}
\expandafter\ifx\csname urlstyle\endcsname\relax
  \providecommand{\doi}[1]{doi: #1}\else
  \providecommand{\doi}{doi: \begingroup \urlstyle{rm}\Url}\fi

\bibitem[Anthropic(2025)]{claude3.7}
Anthropic.
\newblock Claude 3.7 sonnet and claude code, 2025.
\newblock URL \url{https://www.anthropic.com/claude/sonnet}.

\bibitem[{Bai} et~al.(2025){Bai}, {Chen}, {Liu}, {Wang}, {Ge}, {Song}, {Dang}, {Wang}, {Wang}, {Tang}, {Zhong}, {Zhu}, {Yang}, {Li}, {Wan}, {Wang}, {Ding}, {Fu}, {Xu}, {Ye}, {Zhang}, {Xie}, {Cheng}, {Zhang}, {Yang}, {Xu}, and {Lin}]{qwen2.5-VL}
Shuai {Bai}, Keqin {Chen}, Xuejing {Liu}, Jialin {Wang}, Wenbin {Ge}, Sibo {Song}, Kai {Dang}, Peng {Wang}, Shijie {Wang}, Jun {Tang}, Humen {Zhong}, Yuanzhi {Zhu}, Mingkun {Yang}, Zhaohai {Li}, Jianqiang {Wan}, Pengfei {Wang}, Wei {Ding}, Zheren {Fu}, Yiheng {Xu}, Jiabo {Ye}, Xi~{Zhang}, Tianbao {Xie}, Zesen {Cheng}, Hang {Zhang}, Zhibo {Yang}, Haiyang {Xu}, and Junyang {Lin}.
\newblock {Qwen2.5-VL Technical Report}.
\newblock \emph{arXiv e-prints}, art. arXiv:2502.13923, February 2025.
\newblock \doi{10.48550/arXiv.2502.13923}.

\bibitem[Bill Yuchen~Lin(2024)]{ZebraLogic}
Yejin~Choi Bill Yuchen~Lin, Ronan Le~Bras.
\newblock Zebralogic: Benchmarking the logical reasoning ability of language models, 2024.
\newblock URL \url{https://hf.co/spaces/allenai/ZebraLogicBench-Leaderboard}.

\bibitem[ByteDance(2025{\natexlab{a}})]{Doubao-vision-pro}
ByteDance.
\newblock Doubao-vision-pro, 2025{\natexlab{a}}.

\bibitem[ByteDance(2025{\natexlab{b}})]{doubao1.5pro}
ByteDance.
\newblock Doubao-1.5-pro, 2025{\natexlab{b}}.
\newblock URL \url{https://team.doubao.com/en/special/doubao_1_5_pro}.

\bibitem[Chevalier-Boisvert et~al.(2023)Chevalier-Boisvert, Dai, Towers, de~Lazcano, Willems, Lahlou, Pal, Castro, and Terry]{MinigridMiniworld23}
Maxime Chevalier-Boisvert, Bolun Dai, Mark Towers, Rodrigo de~Lazcano, Lucas Willems, Salem Lahlou, Suman Pal, Pablo~Samuel Castro, and Jordan Terry.
\newblock Minigrid \& miniworld: Modular \& customizable reinforcement learning environments for goal-oriented tasks.
\newblock \emph{CoRR}, abs/2306.13831, 2023.

\bibitem[{Cobbe} et~al.(2021){Cobbe}, {Kosaraju}, {Bavarian}, {Chen}, {Jun}, {Kaiser}, {Plappert}, {Tworek}, {Hilton}, {Nakano}, {Hesse}, and {Schulman}]{GSM8K}
Karl {Cobbe}, Vineet {Kosaraju}, Mohammad {Bavarian}, Mark {Chen}, Heewoo {Jun}, Lukasz {Kaiser}, Matthias {Plappert}, Jerry {Tworek}, Jacob {Hilton}, Reiichiro {Nakano}, Christopher {Hesse}, and John {Schulman}.
\newblock {Training Verifiers to Solve Math Word Problems}.
\newblock \emph{arXiv e-prints}, art. arXiv:2110.14168, October 2021.
\newblock \doi{10.48550/arXiv.2110.14168}.

\bibitem[DeepMind(2024)]{gemini-thinking}
Google DeepMind.
\newblock Gemini 2.0 flash thinking, 2024.
\newblock URL \url{https://deepmind.google/technologies/gemini/flash-thinking/}.

\bibitem[DeepMind(2025)]{gemini2.5}
Google DeepMind.
\newblock Gemini 2.5: Our most intelligent ai model, 2025.
\newblock URL \url{https://blog.google/technology/google-deepmind/gemini-model-thinking-updates-march-2025/}.

\bibitem[Gong et~al.(2024)Gong, Huang, Ma, Noda, Durante, Zheng, Terzopoulos, Fei{-}Fei, Gao, and Vo]{mindagent}
Ran Gong, Qiuyuan Huang, Xiaojian Ma, Yusuke Noda, Zane Durante, Zilong Zheng, Demetri Terzopoulos, Li~Fei{-}Fei, Jianfeng Gao, and Hoi Vo.
\newblock Mindagent: Emergent gaming interaction.
\newblock In Kevin Duh, Helena G{\'{o}}mez{-}Adorno, and Steven Bethard, editors, \emph{Findings of the Association for Computational Linguistics: {NAACL} 2024, Mexico City, Mexico, June 16-21, 2024}, pages 3154--3183. Association for Computational Linguistics, 2024.
\newblock \doi{10.18653/V1/2024.FINDINGS-NAACL.200}.
\newblock URL \url{https://doi.org/10.18653/v1/2024.findings-naacl.200}.

\bibitem[{Guertler} et~al.(2025){Guertler}, {Cheng}, {Yu}, {Liu}, {Choshen}, and {Tan}]{textarena}
Leon {Guertler}, Bobby {Cheng}, Simon {Yu}, Bo~{Liu}, Leshem {Choshen}, and Cheston {Tan}.
\newblock {TextArena}.
\newblock \emph{arXiv e-prints}, art. arXiv:2504.11442, April 2025.
\newblock \doi{10.48550/arXiv.2504.11442}.

\bibitem[{Gui} et~al.(2024){Gui}, {Liu}, {Cheng}, {Gu}, {Liu}, {Wang}, {Dong}, {Tang}, and {Huang}]{logicgame}
Jiayi {Gui}, Yiming {Liu}, Jiale {Cheng}, Xiaotao {Gu}, Xiao {Liu}, Hongning {Wang}, Yuxiao {Dong}, Jie {Tang}, and Minlie {Huang}.
\newblock {LogicGame: Benchmarking Rule-Based Reasoning Abilities of Large Language Models}.
\newblock \emph{arXiv e-prints}, art. arXiv:2408.15778, August 2024.
\newblock \doi{10.48550/arXiv.2408.15778}.

\bibitem[Guo et~al.(2025)Guo, Yang, Zhang, Song, Zhang, Xu, Zhu, Ma, Wang, Bi, et~al.]{guo2025deepseek}
Daya Guo, Dejian Yang, Haowei Zhang, Junxiao Song, Ruoyu Zhang, Runxin Xu, Qihao Zhu, Shirong Ma, Peiyi Wang, Xiao Bi, et~al.
\newblock Deepseek-r1: Incentivizing reasoning capability in llms via reinforcement learning.
\newblock \emph{arXiv preprint arXiv:2501.12948}, 2025.

\bibitem[{Han} et~al.(2022){Han}, {Schoelkopf}, {Zhao}, {Qi}, {Riddell}, {Zhou}, {Coady}, {Peng}, {Qiao}, {Benson}, {Sun}, {Wardle-Solano}, {Szabo}, {Zubova}, {Burtell}, {Fan}, {Liu}, {Wong}, {Sailor}, {Ni}, {Nan}, {Kasai}, {Yu}, {Zhang}, {Fabbri}, {Kryscinski}, {Yavuz}, {Liu}, {Lin}, {Joty}, {Zhou}, {Xiong}, {Ying}, {Cohan}, and {Radev}]{FOLIO}
Simeng {Han}, Hailey {Schoelkopf}, Yilun {Zhao}, Zhenting {Qi}, Martin {Riddell}, Wenfei {Zhou}, James {Coady}, David {Peng}, Yujie {Qiao}, Luke {Benson}, Lucy {Sun}, Alex {Wardle-Solano}, Hannah {Szabo}, Ekaterina {Zubova}, Matthew {Burtell}, Jonathan {Fan}, Yixin {Liu}, Brian {Wong}, Malcolm {Sailor}, Ansong {Ni}, Linyong {Nan}, Jungo {Kasai}, Tao {Yu}, Rui {Zhang}, Alexander~R. {Fabbri}, Wojciech {Kryscinski}, Semih {Yavuz}, Ye~{Liu}, Xi~Victoria {Lin}, Shafiq {Joty}, Yingbo {Zhou}, Caiming {Xiong}, Rex {Ying}, Arman {Cohan}, and Dragomir {Radev}.
\newblock {FOLIO: Natural Language Reasoning with First-Order Logic}.
\newblock \emph{arXiv e-prints}, art. arXiv:2209.00840, September 2022.
\newblock \doi{10.48550/arXiv.2209.00840}.

\bibitem[{Hendrycks} et~al.(2020){Hendrycks}, {Burns}, {Basart}, {Zou}, {Mazeika}, {Song}, and {Steinhardt}]{MMLU}
Dan {Hendrycks}, Collin {Burns}, Steven {Basart}, Andy {Zou}, Mantas {Mazeika}, Dawn {Song}, and Jacob {Steinhardt}.
\newblock {Measuring Massive Multitask Language Understanding}.
\newblock \emph{arXiv e-prints}, art. arXiv:2009.03300, September 2020.
\newblock \doi{10.48550/arXiv.2009.03300}.

\bibitem[Hendrycks et~al.(2021)Hendrycks, Burns, Kadavath, Arora, Basart, Tang, Song, and Steinhardt]{MATH}
Dan Hendrycks, Collin Burns, Saurav Kadavath, Akul Arora, Steven Basart, Eric Tang, Dawn Song, and Jacob Steinhardt.
\newblock Measuring mathematical problem solving with the math dataset.
\newblock \emph{arXiv preprint arXiv:2103.03874}, 2021.

\bibitem[{Hu} et~al.(2024){Hu}, {Li}, {Xie}, {Jiang}, {Stoica}, {Jin}, and {Zhang}]{gamearena}
Lanxiang {Hu}, Qiyu {Li}, Anze {Xie}, Nan {Jiang}, Ion {Stoica}, Haojian {Jin}, and Hao {Zhang}.
\newblock {GameArena: Evaluating LLM Reasoning through Live Computer Games}.
\newblock \emph{arXiv e-prints}, art. arXiv:2412.06394, December 2024.
\newblock \doi{10.48550/arXiv.2412.06394}.

\bibitem[Light et~al.(2023)Light, Cai, Shen, and Hu]{avalonbench}
Jonathan Light, Min Cai, Sheng Shen, and Ziniu Hu.
\newblock Avalonbench: Evaluating llms playing the game of avalon.
\newblock \emph{arXiv preprint arXiv:2310.05036}, 2023.

\bibitem[Liu et~al.(2024)Liu, Feng, Xue, Wang, Wu, Lu, Zhao, Deng, Zhang, Ruan, et~al.]{deepseekv3}
Aixin Liu, Bei Feng, Bing Xue, Bingxuan Wang, Bochao Wu, Chengda Lu, Chenggang Zhao, Chengqi Deng, Chenyu Zhang, Chong Ruan, et~al.
\newblock Deepseek-v3 technical report.
\newblock \emph{arXiv preprint arXiv:2412.19437}, 2024.

\bibitem[{Liu} et~al.(2023){Liu}, {Yu}, {Zhang}, {Xu}, {Lei}, {Lai}, {Gu}, {Ding}, {Men}, {Yang}, {Zhang}, {Deng}, {Zeng}, {Du}, {Zhang}, {Shen}, {Zhang}, {Su}, {Sun}, {Huang}, {Dong}, and {Tang}]{AgentBench}
Xiao {Liu}, Hao {Yu}, Hanchen {Zhang}, Yifan {Xu}, Xuanyu {Lei}, Hanyu {Lai}, Yu~{Gu}, Hangliang {Ding}, Kaiwen {Men}, Kejuan {Yang}, Shudan {Zhang}, Xiang {Deng}, Aohan {Zeng}, Zhengxiao {Du}, Chenhui {Zhang}, Sheng {Shen}, Tianjun {Zhang}, Yu~{Su}, Huan {Sun}, Minlie {Huang}, Yuxiao {Dong}, and Jie {Tang}.
\newblock {AgentBench: Evaluating LLMs as Agents}.
\newblock \emph{arXiv e-prints}, art. arXiv:2308.03688, August 2023.
\newblock \doi{10.48550/arXiv.2308.03688}.

\bibitem[{Ma} et~al.(2024){Ma}, {Du}, {Wang}, {Zhang}, {Wen}, {Qu}, {Yang}, {Liu}, {Liu}, {Yue}, {Huang}, and {Zhang}]{korbench}
Kaijing {Ma}, Xinrun {Du}, Yunran {Wang}, Haoran {Zhang}, Zhoufutu {Wen}, Xingwei {Qu}, Jian {Yang}, Jiaheng {Liu}, Minghao {Liu}, Xiang {Yue}, Wenhao {Huang}, and Ge~{Zhang}.
\newblock {KOR-Bench: Benchmarking Language Models on Knowledge-Orthogonal Reasoning Tasks}.
\newblock \emph{arXiv e-prints}, art. arXiv:2410.06526, October 2024.
\newblock \doi{10.48550/arXiv.2410.06526}.

\bibitem[OpenAI(2024{\natexlab{a}})]{o1}
OpenAI.
\newblock Learning to reason with llms, 2024{\natexlab{a}}.
\newblock URL \url{https://openai.com/index/learning-to-reason-with-llms/}.

\bibitem[OpenAI(2024{\natexlab{b}})]{openai2024gpt4o}
OpenAI.
\newblock Gpt-4o system card.
\newblock Technical report, OpenAI, 2024{\natexlab{b}}.
\newblock \url{https://www.openai.com/research/gpt-4o}.

\bibitem[OpenAI(2025)]{o3}
OpenAI.
\newblock Learning to reason with llms, 2025.
\newblock URL \url{https://openai.com/index/openai-o3-mini/}.

\bibitem[{P Team} et~al.(2025){P Team}, {Du}, {Yao}, {Ma}, {Wang}, {Zheng}, {Zhu}, {Liu}, {Liang}, {Jin}, {Wei}, {Zheng}, {Deng}, {Gavin}, {Jia}, {Jiang}, {Liao}, {Li}, {Li}, {Li}, {Li}, {Li}, {Ma}, {Ni}, {Que}, {Wang}, {Wen}, {Wu}, {Hsing}, {Xu}, {Yang}, {Moore Wang}, {Zhou}, {Bai}, {Bu}, {Cai}, {Chen}, {Chen}, {Cheng}, {Cheng}, {Ding}, {Huang}, {Huang}, {Li}, {Li}, {Li}, {Liang}, {Lin}, {Lin}, {Ma}, {Pang}, {Peng}, {Peng}, {Qi}, {Qiu}, {Qu}, {Quan}, {Tan}, {Wang}, {Wang}, {Wang}, {Wang}, {Wang}, {Xu}, {Yang}, {Yuan}, {Yue}, {Zhan}, {Zhang}, {Zhang}, {Zhang}, {Zhang}, {Zhang}, {Zhao}, {Zheng}, {Zhong}, {Gao}, {Li}, {Liu}, {Liu}, {Liu}, {Ni}, {Peng}, {Qin}, {Su}, {Wang}, {Wang}, {Yang}, {Yang}, {Cao}, {Yue}, {Zhang}, {Zhou}, {Liu}, {Lin}, {Huang}, and {Zhang}]{SuperGPQA}
{P Team}, Xinrun {Du}, Yifan {Yao}, Kaijing {Ma}, Bingli {Wang}, Tianyu {Zheng}, King {Zhu}, Minghao {Liu}, Yiming {Liang}, Xiaolong {Jin}, Zhenlin {Wei}, Chujie {Zheng}, Kaixin {Deng}, Shawn {Gavin}, Shian {Jia}, Sichao {Jiang}, Yiyan {Liao}, Rui {Li}, Qinrui {Li}, Sirun {Li}, Yizhi {Li}, Yunwen {Li}, David {Ma}, Yuansheng {Ni}, Haoran {Que}, Qiyao {Wang}, Zhoufutu {Wen}, Siwei {Wu}, Tyshawn {Hsing}, Ming {Xu}, Zhenzhu {Yang}, Zekun {Moore Wang}, Junting {Zhou}, Yuelin {Bai}, Xingyuan {Bu}, Chenglin {Cai}, Liang {Chen}, Yifan {Chen}, Chengtuo {Cheng}, Tianhao {Cheng}, Keyi {Ding}, Siming {Huang}, Yun {Huang}, Yaoru {Li}, Yizhe {Li}, Zhaoqun {Li}, Tianhao {Liang}, Chengdong {Lin}, Hongquan {Lin}, Yinghao {Ma}, Tianyang {Pang}, Zhongyuan {Peng}, Zifan {Peng}, Qige {Qi}, Shi {Qiu}, Xingwei {Qu}, Shanghaoran {Quan}, Yizhou {Tan}, Zili {Wang}, Chenqing {Wang}, Hao {Wang}, Yiya {Wang}, Yubo {Wang}, Jiajun {Xu}, Kexin {Yang}, Ruibin {Yuan}, Yuanhao {Yue}, Tianyang {Zhan}, Chun {Zhang}, Jinyang {Zhang}, Xiyue
  {Zhang}, Xingjian {Zhang}, Yue {Zhang}, Yongchi {Zhao}, Xiangyu {Zheng}, Chenghua {Zhong}, Yang {Gao}, Zhoujun {Li}, Dayiheng {Liu}, Qian {Liu}, Tianyu {Liu}, Shiwen {Ni}, Junran {Peng}, Yujia {Qin}, Wenbo {Su}, Guoyin {Wang}, Shi {Wang}, Jian {Yang}, Min {Yang}, Meng {Cao}, Xiang {Yue}, Zhaoxiang {Zhang}, Wangchunshu {Zhou}, Jiaheng {Liu}, Qunshu {Lin}, Wenhao {Huang}, and Ge~{Zhang}.
\newblock {SuperGPQA: Scaling LLM Evaluation across 285 Graduate Disciplines}.
\newblock \emph{arXiv e-prints}, art. arXiv:2502.14739, February 2025.
\newblock \doi{10.48550/arXiv.2502.14739}.

\bibitem[{Patel} et~al.(2024){Patel}, {Chakraborty}, {Suttle}, {Wang}, {Singh Bedi}, and {Manocha}]{AIME}
Bhrij {Patel}, Souradip {Chakraborty}, Wesley~A. {Suttle}, Mengdi {Wang}, Amrit {Singh Bedi}, and Dinesh {Manocha}.
\newblock {AIME: AI System Optimization via Multiple LLM Evaluators}.
\newblock \emph{arXiv e-prints}, art. arXiv:2410.03131, October 2024.
\newblock \doi{10.48550/arXiv.2410.03131}.

\bibitem[{Phan} et~al.(2025){Phan}, {Gatti}, {Han}, {Li}, {Hu}, {Zhang}, {Calvin Zhang}, {Shaaban}, {Ling}, {Shi}, {Choi}, {Agrawal}, {Chopra}, {Khoja}, {Kim}, {Ren}, {Hausenloy}, {Zhang}, {Mazeika}, {Dodonov}, {Nguyen}, {Lee}, {Anderson}, {Doroshenko}, {Cennyth Stokes}, {Mahmood}, {Pokutnyi}, {Iskra}, {Wang}, {Levin}, {Kazakov}, {Feng}, {Feng}, {Zhao}, {Yu}, {Gangal}, {Zou}, {Wang}, {Popov}, {Gerbicz}, {Galgon}, {Schmitt}, {Yeadon}, {Lee}, {Sauers}, {Sanchez}, {Giska}, {Roth}, {Riis}, {Utpala}, {Burns}, {Goshu}, {Maheshbhai Naiya}, {Agu}, {Giboney}, {Cheatom}, {Fournier-Facio}, {Crowson}, {Finke}, {Cheng}, {Zampese}, {Hoerr}, {Nandor}, {Park}, {Gehrunger}, {Cai}, {McCarty}, {Garretson}, {Taylor}, {Sileo}, {Ren}, {Qazi}, {Li}, {Nam}, {Wydallis}, {Arkhipov}, {Lun Shi}, {Bacho}, {Willcocks}, {Cao}, {Motwani}, {de Oliveira Santos}, {Veith}, {Vendrow}, {Cojoc}, {Zenitani}, {Robinson}, {Tang}, {Li}, {Vendrow}, {Wildner Fraga}, {Kuchkin}, {Pupasov Maksimov}, {Marion}, {Efremov}, {Lynch}, {Liang}, {Mikov},
  {Gritsevskiy}, {Guillod}, {Demir}, {Martinez}, {Pageler}, {Zhou}, {Soori}, {Press}, {Tang}, {Rissone}, {Green}, {Br{\"u}ssel}, {Twayana}, {Dieuleveut}, {Marvin Imperial}, {Prabhu}, {Yang}, {Crispino}, {Rao}, {Zvonkine}, {Loiseau}, {Kalinin}, {Lukas}, {Manolescu}, {Stambaugh}, {Mishra}, {Hogg}, {Bosio}, {Coppola}, {Salazar}, {Jin}, {Sayous}, {Ivanov}, {Schwaller}, {Senthilkuma}, {Bran}, {Algaba}, {Van den Houte}, {Van Der Sypt}, {Verbeken}, {Noever}, {Kopylov}, {Myklebust}, {Li}, {Schut}, {Zheltonozhskii}, {Yuan}, {Lim}, {Stanley}, {Yang}, {Maar}, {Wykowski}, {Oller}, {Sahu}, {Giulio Ardito}, {Hu}, {Ghislain Kemogne Kamdoum}, {Jin}, {Garcia Vilchis}, {Zu}, {Lackner}, {Koppel}, {Sun}, {Antonenko}, {Chern}, {Zhao}, {Arsene}, {Cavanagh}, {Li}, {Shen}, {Crisostomi}, {Zhang}, {Dehghan}, {Ivanov}, {Perrella}, {Kaparov}, {Zang}, {Sucholutsky}, {Kharlamova}, {Orel}, {Poritski}, {Ben-David}, {Berger}, {Whitfill}, {Foster}, {Munro}, {Ho}, {Sivarajan}, {Bar Hava}, {Kuchkin}, {Holmes}, {Rodriguez-Romero}, {Sommerhage},
  {Zhang}, {Moat}, {Schneider}, {Kazibwe}, {Clarke}, {Kim}, {Meneguitti Dias}, {Fish}, and {Elser}]{HLE}
Long {Phan}, Alice {Gatti}, Ziwen {Han}, Nathaniel {Li}, Josephina {Hu}, Hugh {Zhang}, Chen~Bo {Calvin Zhang}, Mohamed {Shaaban}, John {Ling}, Sean {Shi}, Michael {Choi}, Anish {Agrawal}, Arnav {Chopra}, Adam {Khoja}, Ryan {Kim}, Richard {Ren}, Jason {Hausenloy}, Oliver {Zhang}, Mantas {Mazeika}, Dmitry {Dodonov}, Tung {Nguyen}, Jaeho {Lee}, Daron {Anderson}, Mikhail {Doroshenko}, Alun {Cennyth Stokes}, Mobeen {Mahmood}, Oleksandr {Pokutnyi}, Oleg {Iskra}, Jessica~P. {Wang}, John-Clark {Levin}, Mstyslav {Kazakov}, Fiona {Feng}, Steven~Y. {Feng}, Haoran {Zhao}, Michael {Yu}, Varun {Gangal}, Chelsea {Zou}, Zihan {Wang}, Serguei {Popov}, Robert {Gerbicz}, Geoff {Galgon}, Johannes {Schmitt}, Will {Yeadon}, Yongki {Lee}, Scott {Sauers}, Alvaro {Sanchez}, Fabian {Giska}, Marc {Roth}, S{\o}ren {Riis}, Saiteja {Utpala}, Noah {Burns}, Gashaw~M. {Goshu}, Mohinder {Maheshbhai Naiya}, Chidozie {Agu}, Zachary {Giboney}, Antrell {Cheatom}, Francesco {Fournier-Facio}, Sarah-Jane {Crowson}, Lennart {Finke}, Zerui {Cheng},
  Jennifer {Zampese}, Ryan~G. {Hoerr}, Mark {Nandor}, Hyunwoo {Park}, Tim {Gehrunger}, Jiaqi {Cai}, Ben {McCarty}, Alexis~C {Garretson}, Edwin {Taylor}, Damien {Sileo}, Qiuyu {Ren}, Usman {Qazi}, Lianghui {Li}, Jungbae {Nam}, John~B. {Wydallis}, Pavel {Arkhipov}, Jack~Wei {Lun Shi}, Aras {Bacho}, Chris~G. {Willcocks}, Hangrui {Cao}, Sumeet {Motwani}, Emily {de Oliveira Santos}, Johannes {Veith}, Edward {Vendrow}, Doru {Cojoc}, Kengo {Zenitani}, Joshua {Robinson}, Longke {Tang}, Yuqi {Li}, Joshua {Vendrow}, Natanael {Wildner Fraga}, Vladyslav {Kuchkin}, Andrey {Pupasov Maksimov}, Pierre {Marion}, Denis {Efremov}, Jayson {Lynch}, Kaiqu {Liang}, Aleksandar {Mikov}, Andrew {Gritsevskiy}, Julien {Guillod}, G{\"o}zdenur {Demir}, Dakotah {Martinez}, Ben {Pageler}, Kevin {Zhou}, Saeed {Soori}, Ori {Press}, Henry {Tang}, Paolo {Rissone}, Sean~R. {Green}, Lina {Br{\"u}ssel}, Moon {Twayana}, Aymeric {Dieuleveut}, Joseph {Marvin Imperial}, Ameya {Prabhu}, Jinzhou {Yang}, Nick {Crispino}, Arun {Rao}, Dimitri {Zvonkine},
  Gabriel {Loiseau}, Mikhail {Kalinin}, Marco {Lukas}, Ciprian {Manolescu}, Nate {Stambaugh}, Subrata {Mishra}, Tad {Hogg}, Carlo {Bosio}, Brian~P {Coppola}, Julian {Salazar}, Jaehyeok {Jin}, Rafael {Sayous}, Stefan {Ivanov}, Philippe {Schwaller}, Shaipranesh {Senthilkuma}, Andres~M {Bran}, Andres {Algaba}, Kelsey {Van den Houte}, Lynn {Van Der Sypt}, Brecht {Verbeken}, David {Noever}, Alexei {Kopylov}, Benjamin {Myklebust}, Bikun {Li}, Lisa {Schut}, Evgenii {Zheltonozhskii}, Qiaochu {Yuan}, Derek {Lim}, Richard {Stanley}, Tong {Yang}, John {Maar}, Julian {Wykowski}, Mart{\'\i} {Oller}, Anmol {Sahu}, Cesare {Giulio Ardito}, Yuzheng {Hu}, Ariel {Ghislain Kemogne Kamdoum}, Alvin {Jin}, Tobias {Garcia Vilchis}, Yuexuan {Zu}, Martin {Lackner}, James {Koppel}, Gongbo {Sun}, Daniil~S. {Antonenko}, Steffi {Chern}, Bingchen {Zhao}, Pierrot {Arsene}, Joseph~M {Cavanagh}, Daofeng {Li}, Jiawei {Shen}, Donato {Crisostomi}, Wenjin {Zhang}, Ali {Dehghan}, Sergey {Ivanov}, David {Perrella}, Nurdin {Kaparov}, Allen {Zang},
  Ilia {Sucholutsky}, Arina {Kharlamova}, Daniil {Orel}, Vladislav {Poritski}, Shalev {Ben-David}, Zachary {Berger}, Parker {Whitfill}, Michael {Foster}, Daniel {Munro}, Linh {Ho}, Shankar {Sivarajan}, Dan {Bar Hava}, Aleksey {Kuchkin}, David {Holmes}, Alexandra {Rodriguez-Romero}, Frank {Sommerhage}, Anji {Zhang}, Richard {Moat}, Keith {Schneider}, Zakayo {Kazibwe}, Don {Clarke}, Dae~Hyun {Kim}, Felipe {Meneguitti Dias}, Sara {Fish}, and Veit {Elser}.
\newblock {Humanity's Last Exam}.
\newblock \emph{arXiv e-prints}, art. arXiv:2501.14249, January 2025.
\newblock \doi{10.48550/arXiv.2501.14249}.

\bibitem[{Qiu} et~al.(2025){Qiu}, {Guo}, {Song}, {Sun}, {Cai}, {Wei}, {Luo}, {Yin}, {Zhang}, {Hu}, {Wang}, {Tang}, {Chang}, {Liu}, {Zhou}, {Zhang}, {Zhang}, {Liu}, {Li}, {Zhang}, {Jing}, {Yin}, {Ren}, {Fu}, {Wang}, {Tian}, {Lv}, {Man}, {Li}, {Tao}, {Sun}, {Liang}, {Mu}, {Li}, {Zhang}, {Zhang}, {Li}, {Xia}, {Lin}, {Shen}, {Chen}, {Xiong}, {Wang}, {Wang}, {Ni}, {Zhang}, {Cui}, {Shao}, {Cao}, {Luo}, {Zhang}, and {Zhu}]{PHYBench}
Shi {Qiu}, Shaoyang {Guo}, Zhuo-Yang {Song}, Yunbo {Sun}, Zeyu {Cai}, Jiashen {Wei}, Tianyu {Luo}, Yixuan {Yin}, Haoxu {Zhang}, Yi~{Hu}, Chenyang {Wang}, Chencheng {Tang}, Haoling {Chang}, Qi~{Liu}, Ziheng {Zhou}, Tianyu {Zhang}, Jingtian {Zhang}, Zhangyi {Liu}, Minghao {Li}, Yuku {Zhang}, Boxuan {Jing}, Xianqi {Yin}, Yutong {Ren}, Zizhuo {Fu}, Weike {Wang}, Xudong {Tian}, Anqi {Lv}, Laifu {Man}, Jianxiang {Li}, Feiyu {Tao}, Qihua {Sun}, Zhou {Liang}, Yushu {Mu}, Zhongxuan {Li}, Jing-Jun {Zhang}, Shutao {Zhang}, Xiaotian {Li}, Xingqi {Xia}, Jiawei {Lin}, Zheyu {Shen}, Jiahang {Chen}, Qiuhao {Xiong}, Binran {Wang}, Fengyuan {Wang}, Ziyang {Ni}, Bohan {Zhang}, Fan {Cui}, Changkun {Shao}, Qing-Hong {Cao}, Ming-xing {Luo}, Muhan {Zhang}, and Hua~Xing {Zhu}.
\newblock {PHYBench: Holistic Evaluation of Physical Perception and Reasoning in Large Language Models}.
\newblock \emph{arXiv e-prints}, art. arXiv:2504.16074, April 2025.
\newblock \doi{10.48550/arXiv.2504.16074}.

\bibitem[{Seed} et~al.(2025){Seed}, {:}, {Chen}, {Fan}, {Liu}, {Liu}, {Lin}, {Wang}, {Wang}, {Wei}, {Xu}, {Yuan}, {Yue}, {Yan}, {Yu}, {Zuo}, {Zhang}, {Zhu}, {An}, {Bai}, {Bao}, {Bin}, {Chen}, {Chen}, {Chen}, {Chen}, {Chen}, {Chen}, {Chen}, {Chen}, {Chen}, {Chen}, {Chen}, {Chen}, {Chi}, {Dai}, {Dai}, {Dai}, {Dou}, {Du}, {Du}, {Duan}, {Dun}, {Fan}, {Feng}, {Feng}, {Feng}, {Fu}, {Fu}, {Fu}, {Ge}, {Guo}, {Han}, {Han}, {Hao}, {Hao}, {He}, {He}, {He}, {Heng}, {Hong}, {Hou}, {Hu}, {Hu}, {Hu}, {Hua}, {Huang}, {Huang}, {Huang}, {Huang}, {Huang}, {Huang}, {Jia}, {Jia}, {Jia}, {Jiang}, {Jiang}, {Jiang}, {Jiang}, {Jiang}, {Jiao}, {Jin}, {Jin}, {Lai}, {Li}, {Li}, {Li}, {Li}, {Li}, {Wan}, {Wang}, {Li}, {Li}, {Li}, {Li}, {Li}, {Li}, {Li}, {Li}, {Liang}, {Liang}, {Lin}, {Lin}, {Lin}, {Liu}, {Liu}, {Liu}, {Liu}, {Liu}, {Liu}, {Liu}, {Liu}, {Liu}, {Liu}, {Liu}, {Liu}, {Liu}, {Liu}, {Liu}, {Liu}, {Long}, {Lou}, {Lou}, {Luo}, {Luo}, {Lv}, {Lv}, {Ma}, {Ma}, {Ma}, {Ma}, {Ma}, {Ma}, {Ma}, {Mao}, {Min}, {Nan}, {Ning}, {Ou}, {Pan},
  {Pang}, {Peng}, {Peng}, {Qian}, {Qian}, {Qiao}, {Qu}, {Ren}, {Ren}, {Shan}, {Shen}, {Shen}, {Shen}, {Sheng}, {Shi}, {Shi}, {Shi}, {Shuai Cao}, {Song}, {Song}, {Su}, {Sun}, {Sun}, {Sun}, {Wan}, {Wang}, {Wang}, {Wang}, {Wang}, {Wang}, {Wang}, {Wang}, {Wang}, {Wang}, {Wang}, {Wang}, {Wang}, {Wang}, {Wang}, {Wang}, {Wang}, {Wang}, {Wei}, {Wei}, {Wei}, {Wei}, {Wu}, {Wu}, {Wu}, {Wu}, {Wu}, {Wu}, {Wu}, {Wu}, {Wu}, {Xi}, {Xia}, {Xian}, {Xiang}, and {Xiang}]{doubao-thinking-pro}
ByteDance {Seed}, {:}, Jiaze {Chen}, Tiantian {Fan}, Xin {Liu}, Lingjun {Liu}, Zhiqi {Lin}, Mingxuan {Wang}, Chengyi {Wang}, Xiangpeng {Wei}, Wenyuan {Xu}, Yufeng {Yuan}, Yu~{Yue}, Lin {Yan}, Qiying {Yu}, Xiaochen {Zuo}, Chi {Zhang}, Ruofei {Zhu}, Zhecheng {An}, Zhihao {Bai}, Yu~{Bao}, Xingyan {Bin}, Jiangjie {Chen}, Feng {Chen}, Hongmin {Chen}, Riwei {Chen}, Liangqiang {Chen}, Zixin {Chen}, Jinsong {Chen}, Siyan {Chen}, Kaiyuan {Chen}, Zhi {Chen}, Jin {Chen}, Jiecao {Chen}, Jinxin {Chi}, Weinan {Dai}, Ning {Dai}, Jiahui {Dai}, Shihan {Dou}, Yantao {Du}, Zhengyin {Du}, Jianhui {Duan}, Chen {Dun}, Ting-Han {Fan}, Jiazhan {Feng}, Junda {Feng}, Ziyuan {Feng}, Yuwei {Fu}, Wenqi {Fu}, Hanjie {Fu}, Hao {Ge}, Hongyi {Guo}, Mingji {Han}, Li~{Han}, Wenhao {Hao}, Xintong {Hao}, Qianyu {He}, Jerry {He}, Feng {He}, Wen {Heng}, Zehua {Hong}, Qi~{Hou}, Liang {Hu}, Shengding {Hu}, Nan {Hu}, Kai {Hua}, Qi~{Huang}, Ziyue {Huang}, Hongzhi {Huang}, Zihao {Huang}, Ting {Huang}, Wenhao {Huang}, Wei {Jia}, Bin {Jia}, Xiaoying
  {Jia}, Yuhua {Jiang}, Haobin {Jiang}, Ziheng {Jiang}, Kaihua {Jiang}, Chengquan {Jiang}, Jianpeng {Jiao}, Xiaoran {Jin}, Xing {Jin}, Xunhao {Lai}, Zheng {Li}, Xiang {Li}, Liyi {Li}, Hongkai {Li}, Zheng {Li}, Shengxian {Wan}, Ya~{Wang}, Yunshui {Li}, Chenggang {Li}, Niuniu {Li}, Siyu {Li}, Xi~{Li}, Xiao {Li}, Aoyan {Li}, Yuntao {Li}, Nianning {Liang}, Xinnian {Liang}, Haibin {Lin}, Weijian {Lin}, Ye~{Lin}, Zhicheng {Liu}, Guanlin {Liu}, Guanlin {Liu}, Chenxiao {Liu}, Yan {Liu}, Gaohong {Liu}, Juncai {Liu}, Chundian {Liu}, Deyi {Liu}, Kaibo {Liu}, Siyao {Liu}, Qi~{Liu}, Yongfei {Liu}, Kang {Liu}, Gan {Liu}, Boyi {Liu}, Rui {Long}, Weiqiang {Lou}, Chenwei {Lou}, Xiang {Luo}, Yao {Luo}, Caiping {Lv}, Heyang {Lv}, Bole {Ma}, Qianli {Ma}, Hongzhi {Ma}, Yiyuan {Ma}, Jin {Ma}, Wenchang {Ma}, Tingting {Ma}, Chen {Mao}, Qiyang {Min}, Zhe {Nan}, Guanghan {Ning}, Jinxiang {Ou}, Haojie {Pan}, Renming {Pang}, Yanghua {Peng}, Tao {Peng}, Lihua {Qian}, Lihua {Qian}, Mu~{Qiao}, Meng {Qu}, Cheng {Ren}, Hongbin {Ren}, Yong
  {Shan}, Wei {Shen}, Ke~{Shen}, Kai {Shen}, Guangming {Sheng}, Jinlong {Shi}, Wenlei {Shi}, Guang {Shi}, Shuai {Shuai Cao}, Yuxin {Song}, Zuquan {Song}, Jing {Su}, Yifan {Sun}, Tao {Sun}, Zewei {Sun}, Borui {Wan}, Zihan {Wang}, Xiaohui {Wang}, Xi~{Wang}, Shuguang {Wang}, Jun {Wang}, Qinlong {Wang}, Chenyuan {Wang}, Shuai {Wang}, Zihan {Wang}, Changbao {Wang}, Jiaqiang {Wang}, Shihang {Wang}, Xuwu {Wang}, Zaiyuan {Wang}, Yuxuan {Wang}, Wenqi {Wang}, Taiqing {Wang}, Chengzhi {Wei}, Houmin {Wei}, Ziyun {Wei}, Shufa {Wei}, Zheng {Wu}, Yonghui {Wu}, Yangjun {Wu}, Bohong {Wu}, Shuang {Wu}, Jingqiao {Wu}, Ning {Wu}, Shuangzhi {Wu}, Jianmin {Wu}, Chenguang {Xi}, Fan {Xia}, Yuqiao {Xian}, Liang {Xiang}, and Boren {Xiang}.
\newblock {Seed-Thinking-v1.5: Advancing Superb Reasoning Models with Reinforcement Learning}.
\newblock \emph{arXiv e-prints}, art. arXiv:2504.13914, April 2025.
\newblock \doi{10.48550/arXiv.2504.13914}.

\bibitem[Shi et~al.(2025)Shi, Wei, Yang, Wang, Yang, Zhang, Huang, Peng, Yang, and Wen]{shi2025cryptoxcompositionalreasoning}
Jiajun Shi, Chaoren Wei, Liqun Yang, Zekun~Moore Wang, Chenghao Yang, Ge~Zhang, Stephen Huang, Tao Peng, Jian Yang, and Zhoufutu Wen.
\newblock Cryptox : Compositional reasoning evaluation of large language models, 2025.
\newblock URL \url{https://arxiv.org/abs/2502.07813}.

\bibitem[{Talmor} et~al.(2018){Talmor}, {Herzig}, {Lourie}, and {Berant}]{CommonsenseQA}
Alon {Talmor}, Jonathan {Herzig}, Nicholas {Lourie}, and Jonathan {Berant}.
\newblock {CommonsenseQA: A Question Answering Challenge Targeting Commonsense Knowledge}.
\newblock \emph{arXiv e-prints}, art. arXiv:1811.00937, November 2018.
\newblock \doi{10.48550/arXiv.1811.00937}.

\bibitem[Team()]{reasoninggym}
Open-Thought Team.
\newblock Reasonggym.
\newblock URL \url{https://github.com/open-thought/reasoning-gym}.

\bibitem[Team(2024)]{qwen2.5}
Qwen Team.
\newblock Qwen2.5: A party of foundation models, September 2024.
\newblock URL \url{https://qwenlm.github.io/blog/qwen2.5/}.

\bibitem[Team(2025{\natexlab{a}})]{qwen3}
Qwen Team.
\newblock Qwen3, 2025{\natexlab{a}}.
\newblock URL \url{https://qwen3.org/}.

\bibitem[Team(2025{\natexlab{b}})]{qwenqwq}
Qwen Team.
\newblock Qwen-qwq, 2025{\natexlab{b}}.
\newblock URL \url{https://qwenlm.github.io/zh/blog/qwq-32b/}.

\bibitem[{Towers} et~al.(2024){Towers}, {Kwiatkowski}, {Terry}, {Balis}, {De Cola}, {Deleu}, {Goul{\~a}o}, {Kallinteris}, {Krimmel}, {KG}, {Perez-Vicente}, {Pierr{\'e}}, {Schulhoff}, {Jet Tai}, {Tan}, and {Younis}]{gym}
Mark {Towers}, Ariel {Kwiatkowski}, Jordan {Terry}, John~U. {Balis}, Gianluca {De Cola}, Tristan {Deleu}, Manuel {Goul{\~a}o}, Andreas {Kallinteris}, Markus {Krimmel}, Arjun {KG}, Rodrigo {Perez-Vicente}, Andrea {Pierr{\'e}}, Sander {Schulhoff}, Jun {Jet Tai}, Hannah {Tan}, and Omar~G. {Younis}.
\newblock {Gymnasium: A Standard Interface for Reinforcement Learning Environments}.
\newblock \emph{arXiv e-prints}, art. arXiv:2407.17032, July 2024.
\newblock \doi{10.48550/arXiv.2407.17032}.

\bibitem[{Verma} et~al.(2025){Verma}, {Huang}, {Chen}, {Klein}, and {Tomlin}]{ggbench}
Vivek {Verma}, David {Huang}, William {Chen}, Dan {Klein}, and Nicholas {Tomlin}.
\newblock {Measuring General Intelligence with Generated Games}.
\newblock \emph{arXiv e-prints}, art. arXiv:2505.07215, May 2025.

\bibitem[Xiao et~al.(2023)Xiao, Xu, Zhang, Wang, and Xia]{comprehension}
Changrong Xiao, Sean~Xin Xu, Kunpeng Zhang, Yufang Wang, and Lei Xia.
\newblock Evaluating reading comprehension exercises generated by {LLM}s: A showcase of {C}hat{GPT} in education applications.
\newblock In Ekaterina Kochmar, Jill Burstein, Andrea Horbach, Ronja Laarmann-Quante, Nitin Madnani, Ana{\"i}s Tack, Victoria Yaneva, Zheng Yuan, and Torsten Zesch, editors, \emph{Proceedings of the 18th Workshop on Innovative Use of NLP for Building Educational Applications (BEA 2023)}, pages 610--625, Toronto, Canada, July 2023. Association for Computational Linguistics.
\newblock \doi{10.18653/v1/2023.bea-1.52}.
\newblock URL \url{https://aclanthology.org/2023.bea-1.52/}.

\bibitem[Xie et~al.(2024)Xie, Zhang, Chen, Zhu, Lou, Tian, Xiao, and Su]{travelplanner}
Jian Xie, Kai Zhang, Jiangjie Chen, Tinghui Zhu, Renze Lou, Yuandong Tian, Yanghua Xiao, and Yu~Su.
\newblock Travelplanner: A benchmark for real-world planning with language agents.
\newblock \emph{arXiv preprint arXiv:2402.01622}, 2024.

\bibitem[Xu et~al.(2023)Xu, Wang, Li, Luo, Wang, Liu, and Liu]{werewolf}
Yuzhuang Xu, Shuo Wang, Peng Li, Fuwen Luo, Xiaolong Wang, Weidong Liu, and Yang Liu.
\newblock Exploring large language models for communication games: An empirical study on werewolf.
\newblock \emph{arXiv preprint arXiv:2309.04658}, 2023.

\bibitem[{Yao} et~al.(2025){Yao}, {Wang}, {Hsieh}, {Zhou}, {Zou}, {Cheng}, {Wang}, and {Viswanath}]{SpinBench}
Jianzhu {Yao}, Kevin {Wang}, Ryan {Hsieh}, Haisu {Zhou}, Tianqing {Zou}, Zerui {Cheng}, Zhangyang {Wang}, and Pramod {Viswanath}.
\newblock {SPIN-Bench: How Well Do LLMs Plan Strategically and Reason Socially?}
\newblock \emph{arXiv e-prints}, art. arXiv:2503.12349, March 2025.
\newblock \doi{10.48550/arXiv.2503.12349}.

\bibitem[Yao et~al.(2025)Yao, Wang, Hsieh, Zhou, Zou, Cheng, Wang, and Viswanath]{spin_bench}
Jianzhu Yao, Kevin Wang, Ryan Hsieh, Haisu Zhou, Tianqing Zou, Zerui Cheng, Zhangyang Wang, and Pramod Viswanath.
\newblock Spin-bench: How well do llms plan strategically and reason socially?
\newblock \emph{arXiv preprint arXiv:2503.12349}, 2025.

\bibitem[{Zhu} et~al.(2025){Zhu}, {Wang}, {Chen}, {Liu}, {Ye}, {Gu}, {Tian}, {Duan}, {Su}, {Shao}, {Gao}, {Cui}, {Wang}, {Cao}, {Liu}, {Wei}, {Zhang}, {Wang}, {Xu}, {Li}, {Wang}, {Deng}, {Li}, {He}, {Jiang}, {Luo}, {Wang}, {He}, {Shi}, {Zhang}, {Shao}, {He}, {Xiong}, {Qu}, {Sun}, {Jiao}, {Lv}, {Wu}, {Zhang}, {Deng}, {Ge}, {Chen}, {Wang}, {Dou}, {Lu}, {Zhu}, {Lu}, {Lin}, {Qiao}, {Dai}, and {Wang}]{internvl3}
Jinguo {Zhu}, Weiyun {Wang}, Zhe {Chen}, Zhaoyang {Liu}, Shenglong {Ye}, Lixin {Gu}, Hao {Tian}, Yuchen {Duan}, Weijie {Su}, Jie {Shao}, Zhangwei {Gao}, Erfei {Cui}, Xuehui {Wang}, Yue {Cao}, Yangzhou {Liu}, Xingguang {Wei}, Hongjie {Zhang}, Haomin {Wang}, Weiye {Xu}, Hao {Li}, Jiahao {Wang}, Nianchen {Deng}, Songze {Li}, Yinan {He}, Tan {Jiang}, Jiapeng {Luo}, Yi~{Wang}, Conghui {He}, Botian {Shi}, Xingcheng {Zhang}, Wenqi {Shao}, Junjun {He}, Yingtong {Xiong}, Wenwen {Qu}, Peng {Sun}, Penglong {Jiao}, Han {Lv}, Lijun {Wu}, Kaipeng {Zhang}, Huipeng {Deng}, Jiaye {Ge}, Kai {Chen}, Limin {Wang}, Min {Dou}, Lewei {Lu}, Xizhou {Zhu}, Tong {Lu}, Dahua {Lin}, Yu~{Qiao}, Jifeng {Dai}, and Wenhai {Wang}.
\newblock {InternVL3: Exploring Advanced Training and Test-Time Recipes for Open-Source Multimodal Models}.
\newblock \emph{arXiv e-prints}, art. arXiv:2504.10479, April 2025.
\newblock \doi{10.48550/arXiv.2504.10479}.

\end{thebibliography}
